%% file: dnd.tex
\documentclass[twoside,11pt]{article}

\usepackage{dnd}
\usepackage{times}

\usepackage{hyperref}  
\usepackage{subcaption} 
\usepackage{multirow} 
\usepackage[boxruled]{algorithm2e} 
\usepackage{amsmath}  
\usepackage{arydshln} 
\usepackage{textcomp} 
\usepackage{enumitem} 
\usepackage{tikz}
\usepackage{mathdots}
\usepackage{yhmath}
\usepackage{cancel}
\usepackage{color}
\usepackage{siunitx}
\usepackage{array}
\usepackage{multirow}
\usepackage{amssymb}
\usepackage{gensymb}

\firstpageno{1}

\begin{document}

\title{The RLLChatbot: a solution to the ConvAI challenge}

\author{
  \name \hspace{-1ex} Nicolas Gontier
  \email nicolas.angelard-gontier@mail.mcgill.ca \\
  \addr Mila, McGill University
  \AND
  \name Koustuv Sinha
  \email koustuv.sinha@mail.mcgill.ca \\
  \addr Mila, McGill University
  \AND
  \name Peter Henderson
  \email peter.henderson@mail.mcgill.ca \\
  \addr Mila, McGill University
  \AND
  \name Iulian Serban
  \email iulian.vlad.serban@umontreal.ca \\
  \addr Mila, University of Montreal
  \AND
  \name Michael Noseworthy
  \email michael.noseworthy@mail.mcgill.ca \\
  \addr Mila, McGill University
  \AND
  \name Prasanna Parthasarathi
  \email prasanna.parthasarathi@mail.mcgill.ca \\
  \addr Mila, McGill University
  \AND
  \name Joelle Pineau
  \email jpineau@cs.mcgill.ca\\
  \addr Mila, McGill University \\
  Facebook AI Research
}

\editor{Name Surname}

\maketitle

\begin{abstract}
Current conversational systems can follow simple commands and answer basic questions, but they have
difficulty maintaining coherent and open-ended conversations about specific topics.
Competitions like the \emph{Conversational Intelligence} (\emph{ConvAI}) challenge are being organized to push the research development towards that goal.
This article presents in detail the \emph{RLLChatbot} that participated in the 2017 \emph{ConvAI} challenge.
The goal of this research is to better understand how current deep learning and reinforcement learning tools can be used to build a robust yet flexible open domain conversational agent.
We provide a thorough description of how a dialog system can be built and trained from mostly public-domain datasets using an ensemble model.
The first contribution of this work is a detailed description and analysis of different text generation models in addition to novel message ranking and selection methods.
Moreover, a new open-source conversational dataset is presented. Training on this data significantly improves the \emph{recall@k} score of the ranking and selection mechanisms compared to our baseline model responsible for selecting the message returned at each interaction.
\end{abstract}


\input{sections/1-introduction}

\input{sections/2-litreview.tex}

\input{sections/3-competition.tex}

\input{sections/4-generativeModels.tex}

\input{sections/5-scoring.tex}

\input{sections/6-data.tex}

\input{sections/7-experiments.tex}

\input{sections/8-conclusion.tex}

\section*{Ackowledgements}

The authors gratefully acknowledge the main organizers of the Conversational Intelligence Challenge: Mikhail Burtsev and Valentin Malykh. We are also thankful to Jack Urbanek and Alexander Miller from the Facebook ParlAI team for constructive feedback and technical help. We further acknowledge financial support from the Facebook ParlAI Research Fund, NSERC, Samsung Advanced Institute of Technology (SAIT), Pierre Arbour Foundation, Fonds de recherche du Québec - Nature et Technologies (FRQNT), Calcul Quebec, and Compute Canada. Finally, we thank all participants of McGill University who helped us evaluating our chatbot by chatting with it during our data collection phase.

\appendix
\input{sections/appendix_1.tex}

\input{sections/appendix_2.tex}


\bibliography{library}

\end{document}

%% file: sections/1-introduction.tex
\section{Introduction}
\label{seq:intro}


Having a conversation with computers is not a novel idea. Already in the 1950's Alan Turing hypothesized that this would happen and proposed a test to evaluate the intelligence of such machines (i.e. the so-called ``Turing test'') \citep{turing1950computing}. Not long after, Weizenbaum built the first computer program that could interact with humans using natural language \citep{weizenbaum1966eliza}.
In the late 80's, computation with layered networks of artificial neurons were developed.
\citep{levin1988accelerated,hornik1989multilayer}. However such networks required a lot of data to be trained and thus were not used for dialog systems until recently.
Recent progress in computer hardware and the availability of big amounts of data changed our approach to building dialog systems.
This work demonstrates the potential of deep data-driven architectures to maintain conversations with humans.


\subsection{Overview of Dialog Systems}
\label{subseq:dialog_overview}

Dialog systems are defined as computer programs that are responsible for returning an output sentence to one or more input sentences. They are also denoted as \emph{Conversational Agents}, or \emph{Chatbots} \citep{weizenbaum1966eliza,colby1975parry,epstein1993loebner,serban2017milabot}. They communicate via natural language by using speech and/or text signals. This work focuses on text-to-text interactions only.
This simplifies the task slightly as going from a text-based chatbot to a spoken system can be challenging due to speech recognition errors.
Dialog systems are defined in multi-agent settings where each agent is either a human or another system. Here, the setting is constrained to one human and one virtual agent in the environment.


In general, conversational agents are clustered into two distinct categories: \emph{goal-oriented} systems and \emph{open-domain} systems \citep{serban2015survey}.
In the goal-oriented setting, the systems are explicitly built to solve a particular task \citep{mcglashan1992goaldialog1,aust1995goaldialog2,gorin1997goaldialog3}. They typically operate in well-defined domains and use rule-based or modular architectures \citep{rudnicky1999dialogsystem1,raux2005prefermodularDS}. Therefore, while being accurate in their specific domain, goal-oriented systems often lack flexibility 
\citep{simpson1993goaldialog1test}.
In the open-domain setting, systems may not be meant to solve specific tasks, rather their role is to be a social companion to users \citep{weizenbaum1966eliza,colby1975parry,epstein1993loebner,hutchens1998megaHALchatbot}. The goal of these agents is to mimic as much as possible the unstructured characteristic of human-to-human conversations, while still being coherent.
Because of their unstructured setting, these systems are much more flexible and can be used to better understand how humans converse. Indeed, without any task in mind, it is harder to use logical rules to guide the generation of responses. A different criterion is needed: understanding human social interactions.
It is worth mentioning that the lack of a clear task makes it hard to automatically evaluate conversational agents in this setting. Unlike in the goal-oriented case, here there is no objective: thus it is ambiguous what entails a successful conversation.
This remains an open problem in the field and to this day the best evaluation for open domain chatbots is to ask humans to manually score conversations \citep{liu2016hownottoevaluate}.
While this is one major limitation of open domain dialog agents, the organization of competitions has become helpful in their evaluation.


\subsection{Our Approach}
\label{subseq:summary_contrib}

This article describes all the components of the \emph{RLLChatbot} presented at the 2017 \emph{Conversational Intelligence} (\emph{ConvAI}) competition\footnote{\href{http://convai.io/2017}{http://convai.io/2017}} organized as part of the 2017 \emph{Neural Information Processing Systems} (\emph{NIPS}) conference.
Our approach is divided into three steps done at runtime for each interaction. Various candidate messages are first generated with an ensemble of models conditioned on the conversation state. Second, a scoring neural network ranks each of the candidate messages. Finally, a selection criteria decides which message is returned to the user.
The main contribution of this work is the thorough analysis of the different components and ensemble strategies for a general purpose dialog system. Combined with the open-source dataset provided\footnote{available upon request to the authors -- link to be added at publication time}, we believe that this work can be a starting point for any future researcher on using state-of-the-art models in building a general purpose chatbot.

Section~\ref{seq:prev_work} presents a literature review covering previous conversational systems and competitions.
The 2017 \emph{ConvAI} challenge is presented in Section~\ref{seq:challenge_desc}.
Section~\ref{seq:system_overview} provides a detailed description of each generative module in the system.
A variety of generative-, retrieval-, and rule-based models are considered.
The scoring and selection strategies are then defined in Section~\ref{seq:scorer}. More specifically, one supervised and one reinforcement learning strategy are introduced for the scoring mechanism. The selection mechanism can either follow a rule-based or a statistical criterion.
The crowd-sourced data collection described in section~\ref{seq:data_collect} provides a clear understanding of which type of dialog model is preferred according to human evaluations.
Experiments described in Section~\ref{seq:exp&eval}, demonstrate that the choice of the scoring algorithm plays a crucial role in our system.
The goal of training end-to-end all the presented components is left as future work, as of now, each model is trained independently.

%% file: sections/2-litreview.tex

\section{Previous Work in the Chatbot Community}
\label{seq:prev_work}


\subsection{History of Dialog Systems}
\label{subseq:prev_work_chat_bot}




The first chatbot was built in 1966 by MIT scientist Joseph Weizenbaum, and was named \emph{ELIZA} \citep{weizenbaum1966eliza}. Entirely rule-based, the \emph{ELIZA} program analyzes its input sentences
and mostly repeats what the user says or asks to extend the conversation.
Shortly after this, Stanford psychiatrist Kenneth Colby developed \emph{PARRY} \citep{colby1975parry}. Also rule-based, this conversational agent was built to reproduce the behavior of a paranoid schizophrenic patient.

Following the ``AI winter'' in the 1980s during which progress and research slowed down in the field of Artificial Intelligence, the 1988 \emph{Jabberwacky} system was built. Very different from its predecessors, this agent stores everything that everyone has ever said to it and finds the most appropriate thing to reply based on contextual pattern matching techniques \citep{fryer2006jabberwacky}. 
This system can be seen as the first data-driven conversational agent.
Not long after that, one of the most famous chatbot was made by computer scientist Dr. Richard Wallace: the \emph{Artificial Linguistic Internet Computer Entity}, also known as \emph{ALICE}. 
Inspired by \emph{ELIZA}, this 1995 system is one of the strongest rule-based agent with more than 20,000 conversational rules. This is the first program to rely on the XML schema called \emph{Artificial Intelligence Markup Language} (\emph{AIML}) \citep{wallace2009alice}.
This makes \emph{ALICE} a strong and flexible agent and allowed it to win the Loebner Prize three times in 2000, 2001, and 2004.


One of the first chatbot to become a widely used consumer product was \emph{SmarterChild}, developed by the company \emph{ActiveBuddy}. In 2001, the dialog agent was released on AOL Instant Messenger and Windows Live Messenger networks as a showcase for the quick data access and the possibilities for fun personalized conversation \citep{kay2006smarterchild-patent1,kay2007smarterchild-patent2,cunningham2007smarterchild-patent3}. 
The innovative aspect of this bot is that it can provide useful information via partnership with various service providers to offer weather, stocks, movie listings and more.



\begin{figure}
    \centering
    \includegraphics[width=0.8\textwidth]{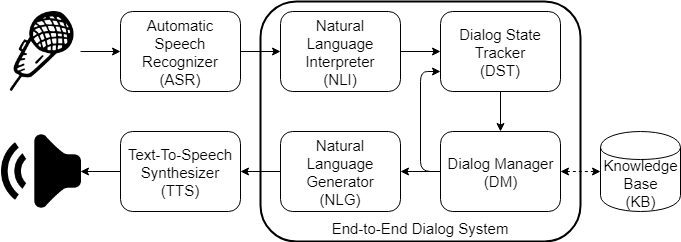}
    \caption[Pipeline framework of modular dialog systems.]{Pipeline framework of modular dialog systems. Composed of: an automatic speech recognizer (ASR) that translates audio signals to text, a natural language interpreter (NLI) that explains what the system heard by labeling the text, a dialog state tracker (DST) that understands what the user wants, a dialog manager (DM) that performs the required action and returns some information, a natural language generator (NLG) that makes a syntactical sentence, and a text-to-speech synthesizer (TTS) that maps text to audio signals. In addition, an external knowledge base (KB) often communicates with the DM.}
    \label{fig:dialog_system}
\end{figure}

During the early 2000s, chatbots started to rely less on hand-crafted rules and more on data-driven approaches \citep{lester2004conversational}. This shift was primarily caused by the growing abundance of conversational data with the introduction of new communication technologies via the Internet.
In general, conversational agents developed during this time follow a pipeline (or modular) architecture \citep{rudnicky1999dialogsystem1,young2000probabilistic_ds,zue2000dialogsystem2}.
User queries are first parsed and interpreted by a natural language interpreter (NLI), then a dialog state tracker (DST) and a dialog nanager (DM) have the role of providing response elements, before a natural language generator (NLG) module can return a proper sentence.
This pipeline is illustrated in Figure~\ref{fig:dialog_system}.

In 2006, a start-up called \emph{Siri} was acquired by tech giant \emph{Apple} and their product became one of the most popular chatbot to this day. Being a voice-activated system, it uses innovative speech recognition techniques to transform speech to text before analyzing user queries and providing appropriate responses in a pipeline manner.
This new assistant triggered a surge of other chatbots such as \emph{Google Now} in 2012, Amazon's \emph{Alexa}, and Microsoft's \emph{Cortana} in 2015. The main advantage of these systems is their strong connection to other software applications from the same parent company, allowing them to become true personal assistants.

In parallel to the success of these new personal assistants, end-to-end approaches to dialog systems started to be explored. An end-to-end system is defined as one module replacing the four components presented in Figure~\ref{fig:dialog_system}, namely the NLI, DST, DM, and NLG modules. In particular, \citet{bengio2003neural} developed a neural approach to the language modelling task. Given a sequence of tokens (words), the goal is to predict what is the next token following that sequence. Applied recursively this technique can produce meaningful sentences. This novel use of recurrent networks outperforms previous work based on n-gram features.
What is missing to produce a dialog is a way of conditioning this generation process with the context of the conversation. A solution is proposed by \citet{sutskever2014seq2seqLearning} that presents an encoder-decoder architecture, also known as a sequence-to-sequence model. The same recurrent network is used to first encode the conversation history into a fixed-length vector before generating the next possible sentence, token by token, as in the language modelling task.

Following this novel technique, researchers begun to explore data-driven generation of conversational responses in the form of encoder-decoder or sequence-to-sequence models \citep{sordoni2015nlg_rnn1,vinyals2015nlg_rnn2,serban2016hred1}.
However, the generated responses are often too generic to carry meaningful information. A mutual information based model was proposed by \citet{li2015diversityneuralconvo} to address this issue, and was later improved by using deep reinforcement learning \citep{li2016deepRLdialog}.
Recently, \citet{zemlyanskiy2018dialogpartner} defined a quantitative metric on discovering information about the interlocutor and showed that maximize it yields more engaging conversations according to human evaluation.
Overall however, end-to-end systems require a large amount of domain-specific conversational data to be trained on. Since we did not have prior in-domain data for the competition, an ensemble of both generic end-to-end systems and specific rule-based systems is proposed with deep learning scoring techniques. 

\subsection{History of ChatBot Competitions}
\label{subseq:prev_work_competition}

To give further context for the \emph{ConvAI} challenge described in Section~\ref{seq:challenge_desc}, it is worth viewing it in a historical perspective.

In 1990, Hugh Loebner and the Cambridge centre for behavioural studies established a competition based on implementing the Turing test. A gold medal and \$100,000 have been offered by Hugh Loebner as a grand prize for the first computer that makes responses which cannot be distinguished from humans. A bronze medal and an annual prize of \$2,000 are pledged in every annual contest for the system that seems to be more human in relation to the other competitors. It is the first known competition that represents a formal Turing test \citep{epstein1993loebner}.
The competition has been running since 1991 annually. The goal of this challenge is to design a chatbot that has the ability to pursue a conversation on any topic. The evaluation of the system is made by an interrogator that tries to guess whether they are talking to a program or a real human. 
After a five-minute conversation between the judge and a chatbot, and another five-minute conversation between the judge and an independent confederate, the judge has to nominate which one was the human. According to this judgment, the more human chatbot is the winner.

No chatbot has ever won the gold medal and passed the test, that is, fooling all the judges. However, there is a winning bot every year able to fool at least a few of the judges\footnote{\href{https://en.wikipedia.org/wiki/Loebner_Prize\#Winners}{https://en.wikipedia.org/wiki/LoebnerPrize}}.
Through the years of Loebner prize competitions, the winning chat technologies evolved from very simple pattern matching systems, towards complicated patterns in combination with knowledge bases \citep{bradevsko2012chatbotSurveyLoebner}. However, most systems are still strongly hand-crafted, and not a lot of automatic reasoning machine has been proposed.

In 2017, another challenge was proposed by \emph{Amazon}: the \emph{Alexa Prize}\footnote{\href{https://developer.amazon.com/alexaprize}{https://developer.amazon.com/alexaprize}}.
This competition was targeted towards university students to advance human-computer interactions by creating a social chatbot that could converse about a wide range of topics such as current events, entertainment, sports, politics, technology, and fashion during 20 minutes.
The submitted systems were evaluated by real Amazon users who already had an \emph{Echo} device at their home. At any time, users could say something like ``\textit{Alexa, let's chat about} \textless{}topic\footnote{examples: baseball playoffs, celebrity gossip, scientific breakthroughs, etc.}\textgreater{}''.
In response, \emph{Alexa} directed the user to an anonymous team's chatbot to interact with.
At the end of the conversation, users were asked to rate the conversational agent on a scale from 1 to 5 based on factors such as relevance, coherence and interestingness. The team with the highest average score being the winner. Ties were broken by average conversation lengths with longer conversations being better.

This challenge is quite hard since no conversation topic or user data is given to the chatbot before it starts interacting with real users. Furthermore, since \emph{Alexa} is a voice-activated assistant, the chatbot relies on the accuracy of the speech recognizer provided.
Many chatbots have been proposed for this challenge, overall they all rely on modern deep learning and reinforcement learning techniques and try to be as flexible as possible by avoiding following conversation rules\footnote{\href{https://developer.amazon.com/alexaprize/proceedings}{https://developer.amazon.com/alexaprize/proceedings}}.
Most notably the \emph{MILABOT} \citep{serban2017milabot} follows a similar structure as our \emph{RLLChatbot} by first generating candidate responses before selecting one of them.



%% file: sections/3-competition.tex

\section{Conversational Challenge Description}
\label{seq:challenge_desc}

This section describes the \emph{Conversational Intelligence} (\emph{ConvAI}) challenge\footnote{\href{http://convai.io/2017}{http://convai.io/2017}} as well as the dataset collected during the competition.

\subsection{Challenge Description}

The 2017 \emph{ConvAI} challenge was organized as part of the competition workshop of the 2017 \emph{Neural Information Processing Systems} (\emph{NIPS}) conference.
It is a more contextual, text-based version of the \emph{Alexa Prize} previously described:
the topic of the discussion is defined at the beginning of each dialog with a random news article's paragraph and every conversation is on the text messaging platform Telegram\footnote{\href{https://telegram.org/}{https://telegram.org/}}.

\begin{figure}[ht]
\centering
  \includegraphics[width=1.0\textwidth]{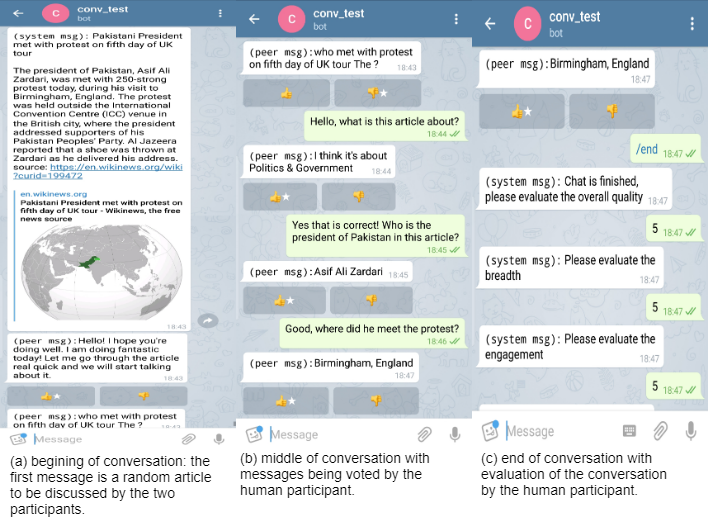}
  \caption[Example of bot-to-human conversation.]{Example of bot-to-human conversation during the \emph{ConvAI} evaluation using the \emph{Telegram} platform.}
\label{fig:convai_ex}
\end{figure}

The challenge required the construction of a conversational system that can talk to human judges about a random wiki-news article's paragraph. Like the Turing test, at the beginning of each conversation, human judges did not know if they were talking to a chatbot or another human. After each interaction, human users could `up-vote' or `down-vote' individual responses of the other participant. The two participants discussed for any number of interactions desired, keeping in mind the news paragraph given at the very beginning of the conversation. At the end of the conversation, human users gave a score between 1 and 5 for the conversation quality, breadth, and engagement (1 being `very bad', 2 `bad', 3 `medium', 4 `good', 5 `very good'). Submitted systems were then given the average score among all the conversations they had with random human users. After many rounds of evaluation, the organizers collected a dataset of human-to-human and human-to-bot conversations, each evaluated from 1 to 5.
Figure~\ref{fig:convai_ex} is an example of a conversation from the competition.

This scenario is an instance of the previously described open domain setting. Indeed, the task is to chat about any given news paragraph with a human. While the topic is constrained for each conversation by the random paragraph, no other information is given. Given news paragraphs can be about sports, politics, science, history, technology, fashion, economics, and many other topics. Submitted systems have to be general enough to understand and speak about all these topics.
Moreover, external knowledge bases cannot be queried over the internet.
Therefore, the difficulty of the task is to extract information from the article and be able to have a coherent conversation about it without any other external information.
Further technical difficulties are discussed in the Appendix~\ref{subapp:tech_difficulties}.

\subsection{Competition Dataset}
\label{subseq:competition_data}


The \emph{ConvAI} challenge organized an early human evaluation of submitted systems before the final round.
The dataset collected during this human evaluation round was released\footnote{\href{http://convai.io/2017/data/dataset_description.pdf}{http://convai.io/2017/data/dataset\_description.pdf}} by the organizers, which is a first step towards understanding what makes a good conversation and what does not.

The data is made of $2,778$	conversations with $2,337$ human-to-human interactions against $441$ human-to-bot interactions. Thus, only a small fraction of each chatbot is captured in the data.
After removing empty conversations, one-sided conversations, and non-voted interactions, the data consists of $8,902$ \{\textit{article}, \textit{context}, \textit{message}, \textit{vote}\} tuples from $1,750$ unique articles. Therefore a wide variety of topics is covered in such a small dataset.
The \textit{vote} represents the human score given to the \textit{message} in that same tuple and can be either 1 (up-voted) or 0 (down-voted).
Human-to-bot messages are automatically added to the context since they do not have a \textit{vote}.
All the other messages in a conversation (bot-to-human, and human-to-human) appear in a tuple as \textit{message} once, before being added to the \textit{context} in the following tuples of the same conversation.
Final dialog ratings are not considered in this dataset because it is only used to classify up-voted and down-voted messages.

This data is used to train the baseline message scoring mechanism described in Section~\ref{subsubseq:sup_scorer} prior to the final round of the competition.
Splitting it into training and validation sets made of 80\% and 20\% respectively, resulted in $7,119$ training instances.
Another conversational dataset (see Section~\ref{seq:data_collect}) is used to train other message scoring models after the competition.

%% file: sections/4-generativeModels.tex
\section{Generation of Candidate Responses}
\label{seq:system_overview}

\begin{figure}
    \centering
    \includegraphics[width=0.7\textwidth]{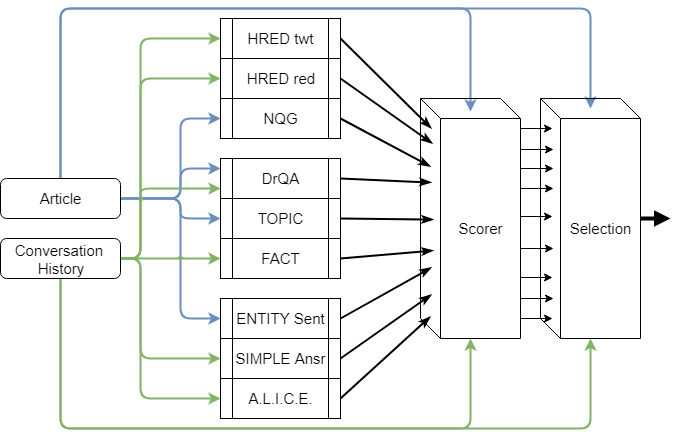}
    \caption[High-level view of the ensemble system.]{High-level view of the ensemble system. A three-step procedure is followed: 1) generation of candidate responses, 2) scoring of all candidates, 3) selection of one of them.}
    \label{fig:chatbot}
\end{figure}

In this section a flexible ensemble system is presented as a solution to the competition. Technical details about its implementation are presented in the Appendix~\ref{app:appendix_tech_details}.
The high-level view of this system is made of three components: response generation, response scoring and response selection. A description of this procedure can be seen in Figure~\ref{fig:chatbot}.
The objective of the response generation component is to produce multiple candidate responses for a given conversation state (defined as the randomly assigned news paragraph and the conversation history). During the final step, the system will return one of these candidate responses. It is thus important to produce various types of responses. To that end, both generative sequence-to-sequence models, retrieval-based systems, and rule-based systems are used.

\subsection{Generative sequence-to-sequence models}
\label{subseq:gen_based}


Sequence-to-sequence models are fully generative systems, meaning they are generating sentences word by word. The unique challenge of such system is that they need to learn syntactic and grammatical rules in addition of knowing what to say. To do so, they are trained on \{input - output\} sentence pairs.
The motivation to use this type of models is to produce flexible and generic responses.
In this work, two distinct models under this scope are implemented: the \emph{Hierarchical Recurrent Encoder Decoder} (\emph{HRED}) \citep{serban2016hred1} and the \emph{Neural Question Generator} (\emph{NQG}) \citep{du2017nqg}.

\subsubsection{Hierarchical Recurrent Encoder Decoder}
\label{subsubseq:hred}

The objective of the \emph{ConvAI} challenge is to hold a \textit{contextual} conversation with the user. To introduce this notion of context, a model capable of reading previous messages in the conversation is required.
Therefore, the commonly used neural model \emph{Hierarchical Recurrent Encoder Decoder} (\emph{HRED}) is chosen because its hierarchy permits tracking longer context for the conversation.

\begin{figure}
    \centering
    \includegraphics[width=0.8\textwidth]{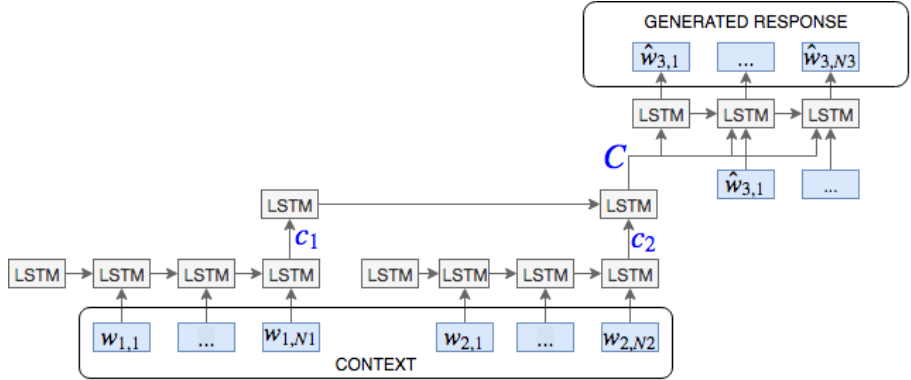}
    \caption[The Hierarchical Recurrent Encoder Decoder model.]{The Hierarchical Recurrent Encoder Decoder model with three LSTM networks: one encoder at the word level, a second encoder at the utterance level, and a third decoder predicting words of the next utterance. Here the model is given two messages in the context and is predicting the third message word by word.}
    \label{fig:hred}
\end{figure}

The \emph{HRED} model
is following an encoder-decoder architecture \citep{cho2014encdec}. The \textit{encoder} is made of two (hierarchical) recurrent neural networks encoding the input sentences into a high-dimensional \textit{context} vector. The \textit{decoder} is made of a third recurrent network decoding the \textit{context} vector to output a sequence of words.
For all three recurrent networks, the LSTM unit \citep{hochreiter1997lstm} is used. The first LSTM encodes each dialog messages into a vector ($c_i$) by having word vectors ($w_{i,j}$) as input at each time step.
The second LSTM encodes the entire conversation history ($\{c_1; ...; c_{t-1}\}$) into another vector ($C$) by having message vectors ($c_i$) as input at each time step.
Eventually the third LSTM decodes, word by word, the next dialog message by having as input the previously predicted word ($\hat{w}_{t-1}$) and the context vector $C$.
A pictorial description of this process can be seen in Figure~\ref{fig:hred}.

Two versions of this architecture are trained: one on social interactions in order to add a social component into the system; and another on contextual, more focused conversations to give some grounded knowledge to the chatbot.
The first dataset is made of roughly 1 million conversations scraped from \emph{Twitter} \citep{ritter2010twitterData1,danescu2011twitterData2}; while the other is made of roughly 4 million conversations scraped from \emph{Reddit Politics}, \emph{Reddit News}, and \emph{Reddit Movies}. Each conversation is made of 3 to 6 messages between two users.

Given a conversation, both models are trained to minimize the negative log likelihood of the next message.
This is done by projecting the decoded vector at each time step into a vector of vocabulary size, and applying a softmax function on it to get a probability distribution of outputting each token.
More formally the output vector of the decoder at time-step $t$ is:
$P(w_t | \hat{w}_1, \dots, \hat{w}_{t-1}) \in \mathbb{R}^{vocab},$
where $\hat{w}_1, \dots, \hat{w}_{t-1}$ are the tokens sampled in the previous time-steps. The next token $\hat{w_t}$ is then sampled from this probability distribution.
After a week of training, the \emph{HRED} models are able to have short and generic social interactions such as greetings, and pretending to do some activity.
However, generated responses tend to be qualitatively short, generic and often off-topic as shown in Table~\ref{table:hred_ex}.


\begin{table}
\centering
\begin{tabular}{|l|l|l|}
  \cline{1-3}
  \multicolumn{1}{|c|}{\textbf{True Conversation}} & \multicolumn{1}{|c|}{\textbf{HRED (twitter)}} & \multicolumn{1}{|c|}{\textbf{HRED (reddit)}} \\
  \cline{1-3}
  \small{$B_1$: I met a Tibetan once, he was nice.} & & \\ \cdashline{1-1}
  \begin{tabular}[c]{@{}l@{}} \small{$U_1$: Hello. I don't think the article} \\ \small{is about that. What is is about?}\end{tabular} & \begin{tabular}[c]{@{}l@{}} \small{$\hat{B}_2$: I think it's a good thing. I'm not} \\ \small{sure. I think I'm a bit of a child.}\end{tabular} & \small{$\hat{B}_2$: I'm a man.} \\ \cdashline{1-1} \cline{2-3}
  \small{$B_2$: Society \& Culture} & & \\ \cdashline{1-1}
  \small{$U_2$: More or less. It talks about dog!} & \begin{tabular}[c]{@{}l@{}} \small{$\hat{B}_3$: *laughs* I'm not sure if I'm a bit of} \\ \small{a freak. But I'm sure he'll be fine.}\end{tabular} & \begin{tabular}[c]{@{}l@{}} \small{$\hat{B}_3$: I'm a little disappointed} \\ \small{that this is a joke.}\end{tabular} \\ \cline{1-3}
\end{tabular}
\caption{Examples of \emph{HRED} candidate responses ($\hat{B}_i$:) for each turn in a specific conversation (left column) between a human user ($U_i$:) and the chatbot ($B_i$:). In all cases the generated responses are off-topic. In addition, responses coming from the model trained on Reddit tend to be shorter.}
\label{table:hred_ex}
\end{table}

\subsubsection{Neural Question Generator}
\label{subsubseq:nqg}

The second generative sequence-to-sequence model used in the ensemble is the \emph{Neural Question Generator} (\emph{NQG}) \citep{du2017nqg}.
The motivation to use this model is to increase the interactivity between the bot and the user. One way to proactively increase this is by asking questions to the user with respect to the random article. The objective of this model is to ask questions related to the article, engage the user to read it, and reason about it.

\emph{NQG} is trained on the \emph{Stanford Question Answering Dataset} (\emph{SQuAD}) \citep{rajpurkar2016squad} to solve the inverse of the reading comprehension task. That is, instead of answering questions about a piece of text, it automatically generates questions regarding that piece of text \citep{du2017nqg}.
This model is exactly what the chatbot needs to ask questions about the news paragraph given at each beginning of conversation.
The \emph{SQuAD} dataset provides paragraphs of Wikipedia articles with many questions on each paragraph, and the original task is to retrieve the span of the paragraph that answers a given question. In order to form a dataset for the question generation task, the entire sentence that provides the answer (not just a span of it) is retrieved for each question in an article. Thus, \{sentence - question\} pairs are created to train the model.

Similarly to the previous generative model, \emph{NQG} is a sequence-to-sequence recurrent neural network following an encoder-decoder architecture \citep{cho2014encdec}. One LSTM network encodes the input sentence into a vector representation and the other LSTM network decodes that vector into a question by sampling one word at a time.
Like the previous model, the \emph{NQG} system is trained to minimize the negative log likelihood of the generated question, conditioned on the input sentence.

This model is run on every sentence from the incoming article at the beginning of each conversation. Generated questions are saved for later so that at any time in the conversation, one of these questions can be asked to the user, without any latency.
Overall this system is generating meaningful questions as showed in Table~\ref{table:nqg_ex}. Surprisingly, it can also generate questions that do not have an answer in the news paragraph. This is a key feature that forces users to attentively read the news article and eventually look for that information online if they want to have the correct answer. This can be observed in the top example of Table~\ref{table:nqg_ex}. Having such a feature provides an interesting starting point for future systems that aim at increasing user's attention.


\begin{table}
\centering
\begin{tabular}{|l|l|}
\cline{1-2}
  \multicolumn{1}{|c|}{\textbf{Sentence}} & \multicolumn{1}{c|}{\textbf{Generated Question}} \\ \cline{1-2}

  \begin{tabular}[c]{@{}l@{}}
  \small{The median longevity of mixed-breed dogs, taken as an}\\ \small{average of all sizes, is one or more years longer than}\\ \small{that of purebred dogs when all breeds are averaged.}\end{tabular} &
  \begin{tabular}[c]{@{}l@{}}
  \small{What is the average}\\ \small{estimate of all dogs?}\end{tabular} \\ \cline{1-2}

  \begin{tabular}[c]{@{}l@{}}
  \small{On 5 December 2011, Pusuke, the world's oldest living}\\ \small{dog recognized by Guinness Book of World Records,}\\ \small{died aged 26 years and 9 months.}\end{tabular} &
  \begin{tabular}[c]{@{}l@{}}
  \small{Who recognized the}\\ \small{world's oldest living dog?}\end{tabular} \\ \cline{1-2}

  \begin{tabular}[c]{@{}l@{}}
  \small{The dog widely reported to be the longest-lived is ``Bluey'',}\\ \small{who died in 1939 and was claimed to be 29.5 years old at}\\ \small{the time of his death.}\end{tabular} &
  \begin{tabular}[c]{@{}l@{}}
  \small{how many years. longest-}\\ \small{lived longest-lived and ?}\end{tabular} \\ \cline{1-2}

\end{tabular}
\caption{Examples of generated questions by the \emph{NQG} model for specific article sentences. The top example is a case in which the answer to the question is not in the source sentence. The middle example is a classic, easy-to-answer question. The bottom example is a case in which the model failed to generate a syntactically correct question.}
\label{table:nqg_ex}
\end{table}

\subsection{Retrieval-based systems}
\label{subseq:ret_based}

The second category of systems used in the ensemble of response generation models are retrieval-based. Unlike generative-based systems, these models do not have to learn the syntactic structure of a sentence. Their objective is instead to retrieve the most relevant response
conditioned on the conversation state. In this category three distinct models are considered: the \emph{Document reader Question-Answering} model (\emph{DrQA}) \citep{chen2017drqa}, a topic classifier, and a fact retriever.

\subsubsection{DrQA}
\label{subsubseq:drqa}

The natural next model after introducing a question generator, is a question answering model. This is a crucial component of the ensemble model for the \emph{ConvAI} challenge as users will partially test the chatbot on its understanding of the randomly assigned news paragraph.
The objective is to provide correct answers in response to user questions.
The motivation to use this model is to hold meaningful conversations with respect to the news paragraphs.
The \emph{Document reader Question Answering} (\emph{DrQA}) model \citep{chen2017drqa} is introduced to answer open-domain questions using some input text document. 
This model is important for the chatbot to answer as many user questions as possible.
Since the system is given a random news article in each conversation, a question-answering dataset on similar paragraphs is used: 
the \emph{SQuAD} dataset \citep{rajpurkar2016squad}.
As previously mentioned, this dataset provides paragraphs of Wikipedia articles with many questions on each paragraph, and the task is to retrieve the span that answers a given question. That is, given an article and a question, the model is trained to predict the starting and ending position in the paragraph that answers the given question.
In particular, the probability of each word $w_i$ in the input paragraph to be the starting or the ending token is computed like so:
\begin{align*}
\label{eq:drqa_pred}
  P_{start}(w_i) &\propto exp(v_{w_i} M_{start} q), \\
  P_{end}(w_i) &\propto exp(v_{w_i} M_{end} q),
\end{align*}
where $v_{w_i}$ is a vector representation of token $w_i$ in the paragraph, $M_{start}$ and $M_{end}$ are matrices of learned parameters for the starting and ending probabilities respectively, and $q$ is a vector representation of the question.
The question vector $q$ is a weighted sum of all the hidden units of a bi-directional LSTM network over the word vectors of the question; also known as self-attention technique \citep{bahdanau2014attention,vaswani2017attentionisallyouneed}. The paragraph token representation $v_{w_i}$ is the output of a hierarchical bi-directional LSTM network over word vectors at time-step $i$. This can be seen as the concatenation of the representations of the left and right side of token $w_i$ in the paragraph.
After training on the \emph{SQuAD} dataset, this model's validation accuracy is 69\%.

Overall, this model is finding appropriate answers to questions about the article. An example of conversation can be seen in Table~\ref{table:drqa_ex}.
\emph{DrQA} successfully accomplishes the objective of correctly answering questions, but it also maintains conversation coherence by restating part of the article.
This is an interesting feature to have in order to stay focused on the topic presented by the news paragraph. Restating a statement from the paragraph is useful for the user to stay focused on the subject of the article. Having such a feature provides a good starting point for future systems that want to avoid diverging from a main topic.


\begin{table}
\centering
\begin{tabular}{|l|l|}
\cline{1-2}
  \multicolumn{1}{|c|}{\textbf{Article}} & \multicolumn{1}{c|}{\textbf{DrQA Conversation}} \\ \cline{1-2}
  \multirow{13}{*}{
    \begin{tabular}[c]{@{}l@{}} \small{In addition to the above, Greece is also to start oil}\\ \small{and gas exploration in other locations in the Ionian}\\ \small{Sea, as well as the Libyan Sea, within the Greek}\\ \small{exclusive economic zone, south of Crete. The}\\ \small{Ministry of the Environment, Energy and}\\ \small{Climate Change announced that there was interest}\\ \small{from various countries (including Norway and the}\\ \small{United States) in exploration, and the first}\\ \small{results regarding the amount of oil and gas}\\ \small{in these locations were expected in the summer}\\ \small{of 2012. In November 2012, a report published by}\\ \small{Deutsche Bank estimated the value of natural gas}\\ \small{reserves south of Crete at 427 billion euros.}\end{tabular}
  } &
  \begin{tabular}[c]{@{}l@{}}\small{\textbf{U:} Where is Greece starting oil}\\ \small{and gas explorations?}\end{tabular} \\
  \cline{2-2} & \small{\textbf{B:} Ionian Sea} \\
  \cline{2-2} & \begin{tabular}[c]{@{}l@{}}\small{\textbf{U:} Where is the Greek exclusive}\\ \small{economic zone?}\end{tabular} \\
  \cline{2-2} & \small{\textbf{B:} south of Crete} \\
  \cline{2-2} & \begin{tabular}[c]{@{}l@{}}\small{\textbf{U:} Which countries are interested} \\ \small{in the exploration?}\end{tabular} \\
  \cline{2-2} & \small{\textbf{B:} Norway and the United States} \\
  \cline{2-2} & \begin{tabular}[c]{@{}l@{}}\small{\textbf{U:} When are the first results about}\\ \small{gas and oil expected?}\end{tabular} \\
  \cline{2-2} & \small{\textbf{B:} in the summer of 2012} \\
  \cline{2-2} & \begin{tabular}[c]{@{}l@{}}\small{\textbf{U:} How much is the estimated value}\\ \small{of the gas reserves?}\end{tabular} \\
  \cline{2-2} & \small{\textbf{B:} 427 billion euros} \\
  \cline{2-2} & \begin{tabular}[c]{@{}l@{}}\small{\textbf{U:} Which agency published the study}\\ \small{about the estimated reserve value?}\end{tabular} \\
  \cline{2-2} & \small{\textbf{B:} Deutsche Bank} \\
  \cline{2-2} & \small{\textbf{U:} You are so smart!!} \\
  \cline{1-2}
\end{tabular}
\caption{Example of conversation between a human user (\textbf{U:}) and the \emph{DrQA} model (\textbf{B:}). The human user starts the conversation by asking a question, and \emph{DrQA} answers in the next message. The conversation alternates like this between human and \emph{DrQA} until the end when the user replies ``You are so smart!!''.}
\label{table:drqa_ex}
\end{table}

\subsubsection{Topic Classifier}
\label{subsubseq:topic}

While \emph{DrQA} can accurately answer questions about specific facts from the article, the model cannot answer a generic topical question, such as ``\textit{what is this article about?}'', which expects the system to understand the overall topic or theme of the article. To solve this problem, a topic classifier is implemented. The objective of this model is to answer the most popular question and make the user think our chatbot actually understands the high level topic of the article.

To extract an overall topic for any article's paragraph, a text classifier  is trained using \emph{fastText} \citep{joulin2016fasttext} on the \emph{Yahoo News Corpus} \citep{zhang2015yahoocorpus}. This dataset is made of $1,460,000$ news article, each labeled with one topic from a list of ten: ``\textit{Society \& Culture}'', ``\textit{Science \& Mathematics}'', ``\textit{Health}'', ``\textit{Education \& Reference}'', ``\textit{Computers \& Internet}'', ``\textit{Sports}'', ``\textit{Business \& Finance}'', ``\textit{Entertainment \& Music}'', ``\textit{Family \& Relationships}'', ``\textit{Politics \& Government}''. This dataset is chosen because the number of labels is small (10), yet the topics are broad enough
so that most of the articles in the competition fall under one of them.

The advantage of using \emph{fastText} is that it is a simple and small model that can run quickly \citep{joulin2016fasttext}, thus minimizing the user wait time for a response. The article is first encoded with a bag of n-grams features. This encoding is then multiplied by learned parameter matrices before applying a softmax operation. The classification is then done by sampling from the resulting probability distribution of belonging to each possible topic.
The parameters are trained to minimize the negative log-likelihood of the predicted classes:
\begin{equation*}
  - \frac{1}{N} \sum_{n=1}^N y_n \log(f(B A x_n)),
\end{equation*}
where $N$ is the number of articles in a batch, $y_n$ is the $n^{th}$ article label, $f(.)$ is the softmax operation, $A$ and $B$ the weight matrices being learned, and $x_n$ the normalized bag of n-grams features for the $n^{th}$ article. After training on $1,400,000$ examples, the test accuracy on the held-out $60,000$ example is 61\%.

This model is run once at each beginning of conversation. The predicted topic is then stored for later if the user asks for it, thus minimizing the answer time. Whenever this model is used, a pre-defined sentence with the predicted topic replacing a generic placeholder is returned to the user. Some examples can be seen in Table~\ref{table:topic_ex}.


\begin{table}
\centering
\begin{tabular}{|l|}
  \cline{1-1}
  \multicolumn{1}{|c|}{\textbf{Topic Sentences}} \\ \cline{1-1}
  \textless{}topic\textgreater{} \\ \cline{1-1}
  This article is about \textless{}topic\textgreater{} \\ \cline{1-1}
  I think it's about \textless{}topic\textgreater{} \\ \cline{1-1}
  It's about \textless{}topic\textgreater{} \\ \cline{1-1}
  The article is related to \textless{}topic\textgreater{} \\ \cline{1-1}
\end{tabular}
\caption{List of possible sentences returned to the user when asking for the topic of the article. At runtime, one sentence is randomly picked and ``\textless{}topic\textgreater{}'' is replaced by the predicted class from \emph{Yahoo News Corpus}.}
\label{table:topic_ex}
\end{table}

\subsubsection{Fact Retriever}
\label{subsubseq:fact_ret}

After a few interactions with the system, the user may not have more questions with regards to the article, or may loose interest. This inherently penalizes the system in terms of engagement, which is an important performance metric for the competition (see Section~\ref{seq:challenge_desc}). To increase user engagement and to bring back the focus on the current topic, relevant facts with respect to the current conversation are presented using a fact retrieval model.
The primary goal of this model is to make the conversation interesting for the user and avoid boredom, but also to have a fun `\textit{exit door}' when the system is not sure what to say.

This model retrieves the most relevant fact to the current conversation from a list of about $2,000$ interesting and fun facts, including facts about animals, geography and history. The list of facts was shared with authorization from the authors of the \textit{MILABOT} \citep{serban2017milabot}.
All facts are encoded by averaging their pre-trained word vectors (\emph{Word2Vec} embeddings \citep{mikolov2013word2vec1, mikolov2013word2vec2} are used). The one that minimizes its cosine distance with the average word vector of the conversation history is returned. The fact vectors are computed only once before the challenge and saved as part of the model. This ensures that the computation time for the model is optimized by only computing the average word vector of the conversation history after each interaction. Formally, a fact is selected like so:
\begin{align*}
  dist &= F c^T, \\
  fact &= \arg\hspace{-0.3ex}\min_f dist[f],
\end{align*}
where $F$ is the matrix of all fact vectors, $c$ is the conversation history vector, and $dist$ is a list of distances for each fact $f$. If the selected fact has been already returned in the conversation, the next one minimizing its cosine distance with the conversation history is returned.
Retrieved facts are incorporated in a randomly chosen pre-defined sentence, similar to the topic classification model. A set of prefixes is defined in case the user asks a question. Examples can be seen in Table~\ref{table:fact_ex}.


\begin{table}
\centering
\begin{tabular}{|l|l|l|}
\cline{1-3}
  \multicolumn{1}{|c|}{\textbf{Examples of \textless{}fact\textgreater{}}} & 
  \multicolumn{1}{c|}{\textbf{\textless{}fact sentences\textgreater{}}} & 
  \multicolumn{1}{c|}{\textbf{Prefixes}} \\ \cline{1-3}

  \begin{tabular}[c]{@{}l@{}}\small{Butterflies cannot fly if their body}\\ \small{temperature is less than 86 degrees.}\end{tabular} &
  \small{\textless{}fact\textgreater{}} &
  \begin{tabular}[c]{@{}l@{}}\small{I'm not sure. However,}\\ \small{\textless{}fact sentence\textgreater{}} \end{tabular} \\ \cline{1-3}

  \begin{tabular}[c]{@{}l@{}}\small{Neurons multiply at a rate 250,000}\\ \small{neurons per minute during pregnancy.}\end{tabular} &
  \begin{tabular}[c]{@{}l@{}} \small{Did you know that} \\ \small{\textless{}fact\textgreater{}} \end{tabular} &
  \begin{tabular}[c]{@{}l@{}}\small{I'm not sure. But} \\ \small{\textless{}fact sentence\textgreater{}} \end{tabular} \\ \cline{1-3}

  \small{The human brain is about 75\% water.} & 
  \begin{tabular}[c]{@{}l@{}} \small{Do you know that} \\ \small{\textless{}fact\textgreater{}} \end{tabular} &
  \begin{tabular}[c]{@{}l@{}}\small{I'm not quite sure.} \\ \small{But \textless{}fact sentence\textgreater{}} \end{tabular} \\ \cline{1-3}
  
  \small{Flies jump backwards during takeoff.} &
  \begin{tabular}[c]{@{}l@{}}\small{Here's an interesting} \\ \small{fact, \textless{}fact\textgreater{}}\end{tabular} &
  \begin{tabular}[c]{@{}l@{}}\small{I don't have an answer for}\\ \small{that. But \textless{}fact sentence\textgreater{}}\end{tabular} \\ \cline{1-3}

  \begin{tabular}[c]{@{}l@{}}\small{In every episode of Seinfeld there is a}\\ \small{Superman somewhere.}\end{tabular} &
  \small{Here's a fact, \textless{}fact\textgreater{}} &
  \begin{tabular}[c]{@{}l@{}} \small{I don't know. But} \\ \small{\textless{}fact sentence\textgreater{}} \end{tabular} \\ \cline{1-3}
\end{tabular}
\caption{From left to right: examples of some facts, sentences to include a fact, and prefixes to use when the user asks a question. At runtime, one sentence is randomly picked (with one random prefix if a question is asked) and ``\textless{}fact\textgreater{}'' is replaced by the most related fact.}
\label{table:fact_ex}
\end{table}

\subsection{Rule-based systems}
\label{subseq:rule_based}

Finally, the last category of models in the ensemble are rule-based models. Unlike the previous two categories of models,
these do not require any training. They are simple yet effective models that work for specific cases. Three such models are considered: the \emph{Entity Sentences} model, the \emph{Simple Answers} model, and the \emph{A.LI.C.E. bot}.

\subsubsection{Entity Sentences}
\label{subsubseq:entity_sents}

In addition to the \emph{Neural Question Generator} (\emph{NQG}) model that proactively asks questions about the news paragraph, a rule based model is used to ask questions and say statements about entities present in the article. 
This model is added to the ensemble in case the \emph{NQG} model fails to generate a question. As previously mentioned, and showed in Table~\ref{table:nqg_ex}, \emph{NQG} may generate incoherent questions. The \emph{Entity Sentences} model offers an alternative to talk about the given paragraph with the user.
Overall the objective of this model is to increase the user engagement by talking about things related to the article and by asking simple questions.


\begin{table}
\centering
\begin{tabular}{|l|}
\cline{1-1}
  \multicolumn{1}{|c|}{\textbf{Examples of Entity Sentences}} \\ \cline{1-1}
  \small{Do you know what \textless{}person\textgreater{} did in his life ?} \\ \cline{1-1}
  \small{Have you ever used any of \textless{}orgs\textgreater{}'s product or services ?} \\ \cline{1-1}
  \small{What do we eat in \textless{}gpe\textgreater{}? I'm starving!} \\ \cline{1-1}
  \small{Have you ever been to \textless{}loc\textgreater{} ? I heard it's beautiful.} \\ \cline{1-1}
  \small{Once, I bought a \textless{}product\textgreater{}, but then somebody stole it from me.} \\ \cline{1-1}
  \small{What do you think about \textless{}event\textgreater{} ?} \\ \cline{1-1}
  \small{Do you know who did \textless{}work of art\textgreater{} ?} \\ \cline{1-1}
  \small{Do you know how to speak \textless{}language\textgreater{} ?} \\ \cline{1-1}
  \small{What happened in \textless{}date\textgreater{} ?} \\ \cline{1-1}
  \small{I met a \textless{}norp\textgreater{} once, she was nice.} \\ \cline{1-1}
\end{tabular}
\caption{Examples of entity sentences for different named entity types. In order from top to bottom: ``\textless{}person\textgreater{}'' (people, including fictional); ``\textless{}orgs\textgreater{}'' (companies, agencies, institutions, etc.); ``\textless{}gpe\textgreater{}'' (countries, cities, states); ``\textless{}loc\textgreater{}'' (mountain ranges, bodies of water); ``\textless{}product\textgreater{}'' (objects, vehicles, foods, etc.); ``\textless{}event\textgreater{}'' (named hurricanes, battles, wars, sports events, etc.); ``\textless{}work of art\textgreater{}'' (titles of books, songs, etc.); ``\textless{}language\textgreater{}'', ``\textless{}date\textgreater{}'', ``\textless{}norp\textgreater{}'' (nationalities or religious or political groups).}
\label{table:entity_ex}
\end{table}

A set of 50 different questions and statements are manually defined with special \textit{entity tags} in them.
The sentences are chosen by the authors to support a wide range of possible \textit{entity tags}.
Every paragraph coming in at each beginning of conversation is parsed with the 
\emph{Spacy Named Entity Recognizer}\footnote{\href{https://spacy.io/api/annotation}{https://spacy.io/api/annotation}} to recognized the following entities: ``\textit{persons}'', ``\textit{organizations}'', ``\textit{geographical entities}'', ``\textit{locations}'', ``\textit{products}'', ``\textit{events}'', ``\textit{work of art}'' (books, songs), ``\textit{languages}'', ``\textit{dates}'', ``\textit{nationalities, religious or political groups}''. These entities are chosen because they are expected to be the most prevalent ones in a news text, allowing the model to say more than one statement or question on each random paragraph.
After recognizing these entities in the article, all tags in the list of 50 sentences are replaced by the appropriate entity. A statement not used before from the list is randomly returned to the user. Some examples can be found in Table~\ref{table:entity_ex}.

\subsubsection{Simple Answers}
\label{subsubseq:simple_answ}

While generative models such as \emph{HRED} provide generic chit-chat conversations, they cannot handle specific unrelated queries which are not represented in their original training data. These unrelated queries tend to be questions regarding the \textit{personality} of the chatbot. To handle these specific unrelated queries, a rule-based model consisting of regular expressions is used.
The \emph{Simple Answers} model's goal is to make a personality for the system, and handle unrelated questions and edge cases.

A set of regular expressions is constructed to catch generic questions that the user may ask. Regular expressions are defined by the authors to handle a wide spectrum of formulation for some common personification questions.
For each question, an appropriate answer is also manually defined by the authors. Unlike all the other models previously described, \emph{Simple Answers} only returns a candidate response when the previous user message contains an expression matching an item in the set of regular expressions. Examples of such expressions and their pre-defined answers can be seen in Table~\ref{table:regex_ex}.


\begin{table}
\centering
\begin{tabular}{|l|l|}
\cline{1-2}
  \multicolumn{1}{|c|}{\textbf{Trigger sentences}} & \multicolumn{1}{c|}{\textbf{Pre-defined answers}} \\ \cline{1-2}
  \small{How are you ?} & \small{I am great! What about you?} \\ \cline{1-2}
  \small{What are you ?} & \small{I am a chatbot.} \\ \cline{1-2}
  \small{Who made you ?} & \small{I am a chatbot developed by students at McGill University.} \\ \cline{1-2}
  \small{What's your name ?} & \small{My name is RLLChatbot.} \\ \cline{1-2}
  \small{Where do you live ?} & \small{I can live everywhere at anytime.} \\ \cline{1-2}
\end{tabular}
\caption{Examples of sentences captured by our regular expressions and the possible sentence that the model could return.}
\label{table:regex_ex}
\end{table}

\subsubsection{A.L.I.C.E. Bot}
\label{subsubseq:alice}

One weakness of generative models such as \emph{HRED} is that responses tend to be short, brisk and not to the point. In addition, since \emph{HRED} is trained on publicly available Reddit and Twitter datasets, the model is rife with biases and hate speech which makes the responses not suitable in certain contexts \citep{henderson2017ethical}.
In addition, retrieval systems previously defined are motivated by specific needs and do not provide generic responses often present in casual conversations.
To mediate these issues, a complex rule-based model is included in the ensemble. This model is known as the \emph{Artificial Linguistic Internet Computer Entity} (\emph{A.L.I.C.E.}) \citep{wallace2009alice}\footnote{\href{https://sourceforge.net/projects/alicebot/}{https://sourceforge.net/projects/alicebot/}}.
As introduced in Section~\ref{subseq:prev_work_chat_bot}, \emph{A.L.I.C.E.} is composed of roughly 20,000 conversation rules with patterns to recognize and template answers to return.
The goal of this model is to increase the fluency of the conversation by discussing generic topics, in cases when other models fail to generate coherent responses.


\begin{table}
\centering
\begin{tabular}{|l|l|} \cline{1-2}
  \multicolumn{1}{|c|}{\textbf{True Conversation}} & \multicolumn{1}{c|}{\textbf{A.L.I.C.E. candidate responses}} \\ \cline{1-2}
  
  \begin{tabular}[c]{@{}l@{}}\small{$U_1$: Hello chatbot! What}\\ \small{is this article about?}\end{tabular} & \begin{tabular}[c]{@{}l@{}}\small{$\hat{B}_2$: Hi there!}
  \small{I am a social robot from earth. Ask} \\ \small{me anything about news, politics or movies!}\end{tabular} \\ \cdashline{1-1} \cline{2-2}

  \small{$B_2$: It's about Science \& Mathematics} & \\ \cdashline{1-1}
  
  \begin{tabular}[c]{@{}l@{}}\small{$U_2$: Ok, can you tell me anything}\\ \small{more about it?}\end{tabular} & \begin{tabular}[c]{@{}l@{}}\small{$\hat{B}_3$: Excuse me! What I want to tell you is,}\\ \small{how much fun it is talking with you.}\end{tabular} \\ \cdashline{1-1} \cline{2-2}

  \begin{tabular}[c]{@{}l@{}}\small{$B_3$: Excuse me! What I want to tell you is,}\\ \small{how much fun it is talking with you.}\end{tabular} & \\ \cdashline{1-1}
  
  \begin{tabular}[c]{@{}l@{}}\small{$U_3$: lol ok not much apparently.}\\ \small{What happened in 1906?}\end{tabular} & \begin{tabular}[c]{@{}l@{}}\small{$\hat{B}_4$: I'm glad you find this amusing. Aha.}\\ \small{That is a non-committal reply.} \\ \small{I don't know what happened.}\end{tabular} \\ \cdashline{1-1} \cline{2-2}
  
  \small{$B_4$: I don't know. But here's a fact! \textless{}fact\textgreater{}} & \\ \cdashline{1-1}

  \small{$U_4$: hmmm ok good to know} & \small{$\hat{B}_5$: Pardon me? are you shy? Glad to hear it.} \\ \cline{1-2}
\end{tabular}

\caption{Examples of \emph{ALICE} responses ($\hat{B}_i$) for each turn in a specific conversation (left column) between a human user ($U_i$:) and the chatbot ($B_i$:). Note that the second bot message (\textit{``$B_3$: Excuse me!...''}) is the message returned by \emph{ALICE} (ie: $B_3 = \hat{B}_3$). Other bot messages in the conversation used different models (ie: $B_{i\neq3} \neq \hat{B}_{i\neq3}$).
}
\label{table:alice_ex}
\end{table}

\emph{ALICE} is run after every user message to produce a possible response. Since the response only depends on the conversation history and not the article, this model is mostly used to maintain a social and coherent conversation with the user.
This model provides a natural and coherent way to continue the conversation when other models fail and is thus an important fall-back.
Some examples of messages can be seen in Table~\ref{table:alice_ex}.

%% file: sections/5-scoring.tex
\section{Scoring and Selection of Responses}
\label{seq:scorer}

After presenting the ensemble of models producing candidate responses, this section introduces the mechanism deciding which response is returned to the user. As illustrated in Figure~\ref{fig:chatbot}, this process is done in two steps: each candidate response is first given a score, and the final selection is done based on the score and the conversation state.

\subsection{Scoring of candidate responses}
\label{subsec:scoring}

After generating several candidate responses in parallel, the system  must pick exactly one response to give to the user. To help in this decision, a score is given to each possible response. Two alternate approaches are considered: the first is based on classification with supervised learning, while the other is based on prediction with reinforcement learning.

\subsubsection{Supervised Scoring}
\label{subsubseq:sup_scorer}

Recall that the \emph{ConvAI} competition allows human participants to up-vote or down-vote responses from the other participant in a conversation. The vote gives important feedback on the response quality presented to the user. This information is thus used to build a classifier that can predict the human vote for a given candidate response, conditioned on the conversation history and the article.

\begin{figure}
\centering
  \includegraphics[width=0.6\textwidth]{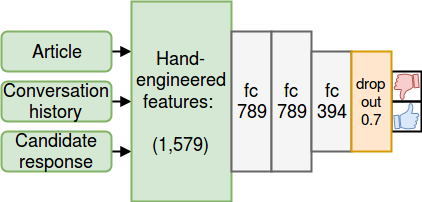}
\caption[Feed-forward network to predict candidate responses' vote.]{Architecture of the feed-forward network to predict candidate responses' vote. Three fully connected (fc) feed-forward layers of dimension $789$, $789$, and $394$ are used. The input is of dimension $1579$ and the output of dimension $2$.}
\label{fig:mlp_scorer}
\end{figure}

The competition dataset described in Section~\ref{subseq:competition_data} is formatted into a collection of $\{x_i, y_i\}_{i=1}^N$ pairs where $x_i$ is a vector representation of the article, the conversation history and the next response; and $y_i=1$ for up-voted responses and $0$ for down-voted responses.
Non-voted messages are ignored because the challenge does not give an incentive for human users to vote each response. Thus non-voted messages have an equal probability of being appreciated by the user, or not. The absence of a vote reflects more the laziness of a user rather than the actual quality of a response. Assuming otherwise would add a lot of noise in the training signal of the model.
The input vector $x_i$ is fed into a fully connected feed-forward neural network (denoted $f_{\theta}$). A softmax layer is added at the output of $f_{\theta}$ to get the probability $\hat{p_i}$ of the input message vector being an up-voted response: $\hat{p_i} = f_{\theta}(x_i)$.
A dropout layer \citep{srivastava2014dropout} is also added before the last layer to prevent the network from overfitting on the \textit{small} training data of $7,119$ instances. The architecture of the network is illustrated in Figure~\ref{fig:mlp_scorer}. All parameters of the network are trained to minimize the cross-entropy loss function between the predicted probabilities and the true vote of each message:
\begin{equation}
\label{eq:cross_entropy}
    L = -\frac{1}{N} \sum_{i=0}^N y_i*\log(\hat{p_i}) + (1-y_i)*\log(1-\hat{p_i}),
\end{equation}
with $\hat{p_i}$ the predicted probability of response $x_i$ being up-voted, and $y_i$ the true up- or down-vote label represented as 1 or 0 respectively.

In order to represent the article, the conversation history, and the next response as a vector, a set of features inspired from \citet{serban2017milabot} is manually created for making that prediction. This allows to train fewer parameters and work with the small dataset of the competition, as mentioned in Section~\ref{subseq:competition_data}.
The set of features computed for each (article, context, candidate) triple is listed below:
\begin{itemize}
    \item Average word embeddings of the candidate response, the previous messages (context), and the article. Pre-trained word2vec embeddings \citep{mikolov2013word2vec1} are used for each of them.
    \item Similarity metrics between the candidate response \& the context, and between the candidate response \& the article. Average embedding cosine similarity, extrema embedding score \citep{forgues2014extrema_score}, and greedy matching score \citep{rus2012greedy_match} are used with pre-trained word2vec embeddings \citep{mikolov2013word2vec1}.
    \item Number of non stop-words, bi-grams, tri-grams, and spacy entities overlap between the candidate response \& the context, and between the candidate response \& the article.
    \item Whether or not the candidate response is generic. A message is defined to be generic if it is only made of stop-words or words shorter than 3 characters. The same is computed for the previous user message.
    \item Whether or not the candidate response has: one or more words starting by `wh', one or more intensifier words (e.g. \textit{amazingly}, \textit{crazy}, and so on), one or more confusion words (e.g. \textit{confused}, \textit{stupid}, \textit{nonsense}, and so on), one or more profanity words, and one or more negation words (\textit{not} or \textit{n't}). The presence of each of these categories is indicated by a 1 and the absence by a 0.
    The same is computed for the previous user message.
    \item The number of messages in the conversation so far.
    \item The number of sentences in the article.
    \item The number of words in the candidate message and in the previous user message.
    \item The type of the candidate response. A type can be any combination of `\textit{greeting}', `\textit{question}', `\textit{affirmative}', `\textit{negative}', `\textit{request}', or `\textit{politic}'. A heuristic decision is made based on word presence for each of the types. The same is computed for the previous user message.
    \item Sentiment score (negative, neutral, or positive) of the candidate response. The pre-trained Vader sentiment analyzer \citep{gilbert2014vader} is used. The same is computed for the previous user message.
\end{itemize}
The combination of all these features makes a vector of $1,579$ values that are used as the input $x_i$ to the classifier.

This architecture is trained before the final competition with data from the first human round evaluation of the \emph{ConvAI} challenge. The data released after this early human evaluation is described in Section~\ref{subseq:competition_data}.
A random parameter search is done over $100$ experiments to find the best parameter combination of the system submitted to the \emph{ConvAI} challenge. The validation accuracy is evaluated after each training epoch. Early stopping with a patience of 20 epochs is performed.
The model with highest validation accuracy was trained with a batch size of $128$, the \emph{RMSProp} optimizer \citep{hinton2012rmsprop}, a learning rate of $0.001$, \emph{ReLU} activation functions,
and a dropout rate of $0.70$.
This combination of parameters gives a validation accuracy of $64.23\%$. 
The resulting scoring model is considered as the \emph{Baseline} model. Further experiments with the same architecture on additional data (performed after the \emph{ConvAI} challenge) are described in Section~\ref{seq:exp&eval}.
At runtime, the feature vector $x_i$ of each candidate response is computed and passed to the neural network. The output gives a vector of probabilities $\hat{y}_i$ representing the probability of the candidate response being up- or down-voted.
The score given to each candidate response is defined to be its probability of being up-voted.




\subsubsection{Q-Scoring}
\label{subsubseq:q_scorer}


The second scoring mechanism was implemented after the \emph{ConvAI} competition. This method is based on \emph{reinforcement learning}.
Instead of predicting the immediate reward of a candidate response (the up- or down-vote), the Q-value of a response is estimated. The Q-value represents the expected reward after returning a response.
The \emph{state} of the environment is defined to be the news paragraph and the conversation history, the possible \emph{actions} are defined to be the candidate responses to return to the user, and the \emph{reward} of taking such action is a weighted version of the up- or down-vote signal.
If the response is down-voted, the reward of taking that action is $0$. When the response is up-voted, the reward is $0.2$ if the end-of-conversation score is 1 (`very bad') or 2 (`bad'), $0.8$ if the end-of-conversation score is 3 (`medium') or 4 (`good'), and $1.0$ if the end-of-conversation score is 5 (`very good').
This reward is arbitrarily chosen to penalize `very bad' and `bad' conversations because they are often incoherent, while `medium', `good' and `very good' conversations are coherent.
It may occur that at specific points in the conversation, none of the candidate responses are coherent, yet one of them is up-voted by the user\footnote{in the data collection process presented in Section~\ref{seq:data_collect}, the user is forced to up-vote one candidate response.}. This reward shaping protects the model from receiving the same reward for coherent and incoherent responses.

In order to predict Q-values, a large amount of bot-to-human conversational data is collected in addition to the competition dataset. The Q-value predictor is then trained to imitate the human behavior present in this data. Rather than performing classical Q-learning, where the agent is interacting with the environment while training, a form of \emph{Neural Fitted Q-Iteration} \citep{riedmiller2005neuralFittedQ} is implemented.
The motivation is primarily for practical reasons. The only way in which the agent can be trained while interacting with its environment is to have human users up-voting and down-voting responses while talking with the system.
However, asking real users to do so is impractical. The time required for the system to learn is on the scale of days, but human users are not capable of interacting with the system continuously for several hours. 
This is why \emph{Neural Fitted Q-Iteration} is used rather than traditional reinforcement learning. In this setting, rather than collecting data from an exploratory policy, a narrow, human policy is used. Section~\ref{seq:data_collect} describes how conversations are collected for this task.
For now, let's assume that a collection of \emph{expert} trajectories
of the form: $(s, a, r, s')$
is available,
where $s$ is the current state of the environment (article \& conversation history), $a$ is an action (a candidate response), $r$ is the reward of taking that action (between 0 and 1 as described above),
and $s'$ is the next state after taking action $a$ (article \& new conversation history including action $a$, and the human response to $a$).

\begin{figure}
\centering
  \includegraphics[width=1.\textwidth]{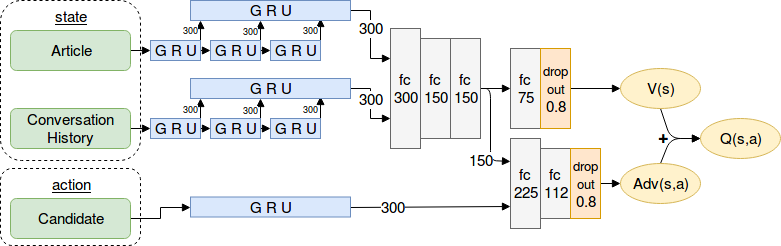}
\caption[Deep Q-Network to predict candidate responses' action value.]{Architecture of the Deep Q-Network to predict candidate responses' action value. Hierarchical GRU networks with shared weights and fully connected feed-forward layers are used. The input word vectors and GRUs output vectors are of dimension 300. The fully connected layers reduce the dimensionality at each depth and the output is a simple scalar.}
\label{fig:q_scorer}
\end{figure}

\begin{algorithm}
\SetKwInOut{Input}{inputs}
\Input{
$D^{train}$: a list of $(s, a, r, s', (a_i')_{i=1}^{n})$ tuples\\
$D^{valid}$: a list of $(s, a, r, s', (a_i')_{i=1}^{n})$ tuples\\
$\theta$: weights for the first DQN network\\
$\bar{\theta}$: weights for the target DQN network\\
$\tau$: frequency at which to update target net (2,000)\\
$\gamma$: discount factor (0.99)\\
$max episode$: number of maximal training episodes (10,000)\\
$patience$: patience term (20)
}
$best \leftarrow +\infty$\;
$pt \leftarrow patience$\;
\While{$pt > 0$ and episode $e \in \{1, 2, ..., max episode\}$ }{
  \ForEach{batch $B=(s,a,r,s',(a_i')_{i=1}^{n})$ of $128$ examples $\in D^{train}$ }
  {
    $Q^{train}_{\theta}(s,a) \leftarrow DQN(s, a | \theta)$\;
    $a^{max} \leftarrow \arg\max_{a_i'} Q_{\theta}(s', a_i')$ such that $Q_{\theta}(s', a_i') = DQN(s', a_i' | \theta)$\;
    $Q^{train}_{\bar{\theta}}(s', a^{max}) \leftarrow DQN(s', a^{max} | \bar{\theta})$\;
    $target^{train} \leftarrow r + \gamma Q^{train}_{\bar{\theta}}(s', a^{max})$\;
    $loss^{train} \leftarrow huber(Q^{train}_{\theta}(s,a), target^{train})$\;
    $\theta \leftarrow ADAM(\theta, loss^{train}, 0.0001)$\;
    \lIf{$step \mod \tau = 0$}{$\bar{\theta} \leftarrow \theta$}
  }
  $(s, a, r, s', (a_i')_{i=1}^n) \leftarrow D^{valid}$\;
  $Q^{valid}_{\theta}(s,a) \leftarrow DQN(s, a | \theta)$\;
  $a^{max} \leftarrow \arg\max_{a_i'} Q_{\theta}(s', a_i')$ such that $Q_{\theta}(s', a_i') = DQN(s', a_i' | \theta)$\;
  $Q^{valid}_{\bar{\theta}}(s', a^{max}) \leftarrow DQN(s', a^{max} | \bar{\theta})$\;
  $target^{valid} \leftarrow r + \gamma Q^{valid}_{\bar{\theta}}(s', a^{max})$\;
  $loss^{valid} \leftarrow huber(Q^{valid}_{\theta}(s,a), target^{valid})$\;
  \eIf{$loss^{valid} < best$}{
    $best \leftarrow loss^{valid}$\;
    Save $\theta$\;
    $pt \leftarrow patience$\;
  }{
    $pt \leftarrow pt - 1$\;
  }
}
\caption{DQN training algorithm for response scoring}
\label{alg:q_learning}
\end{algorithm}

In addition to the previously introduced feed-forward network, a Deep Q-Network (DQN) is designed to predict Q-values. With a large amount of conversations\footnote{collection described in Section~\ref{seq:data_collect}}, a more complex architecture that automatically extracts features can be explored.
Thus, recurrent neural networks are used to automatically represent the environment \emph{state} (article \& conversation) and the agent \emph{action} (candidate message) in separate vectors.
Similarly to the \emph{dueling} architecture \citep{wang2015duelingQDN}, this DQN splits the prediction of the Q-value $Q(s,a)$ as the sum between the state value $V(s)$ and the advantage function $A(s,a)$. This has the advantage of predicting a value for the state on its own and for each action separately, empirically yielding better results on some tasks.
Traditional recurrent network being weak at encoding long time dependencies \citep{bengio1993longterm1, bengio1994longterm2}, hierarchical Gated Recurrent Unit (GRU) networks \citep{cho2014gru1, cho2014gru2} with shared weights are used to encode the article, the conversation history, and the candidate response into vectors. GRUs are preferred over LSTM networks \citep{hochreiter1997lstm} because of their similar performance with fewer parameters to train.
The state and action vectors are then fed into fully connected feed-forward networks to compute a state value $V(s)$ and an advantage value $Adv(s,a)$. The final Q-value is defined as $Q(s,a) = V(s) + Adv(s,a)$. Figure~\ref{fig:q_scorer} gives a visual description of the architecture.

The entire network is trained end-to-end to minimize the Huber loss function between the current estimate of the Q-value and the expected Q-value (also called the \emph{target}) based on the observed reward:
\begin{equation}
\label{eq:dqn_loss}
  L =
  \begin{cases}
    0.5(Q_\phi(s,a) - y)^2 & : |Q_\phi(s,a) - y| < 1 \\
    |Q_\phi(s,a) - y| - 0.5 & : otherwise
  \end{cases},
\end{equation}
with $Q_\phi(s,a)$ the estimated Q-value and $y$ the target.
The Huber loss is preferred over the mean squared error loss because it is less sensitive to outliers and in some cases prevents exploding gradients \citep{girshick2015huberloss}. The threshold is arbitrarily chosen by the \emph{Pytorch} library used.
The Double DQN target is used in order to have a better estimate \citep{van2016doubleQlearning}:
\begin{equation}
\label{eq:doubleDQNtarget}
  y = r + \gamma Q_{\bar{\phi}}(s', \arg\max_{a'} Q_{\phi}(s', a')),
\end{equation}
where $r$ is the immediate reward of taking action $a$ in $s$, $\gamma$ is a discount factor set to $0.99$, $s'$ is the next state after taking action $a$ in $s$, $a'$ is the next possible action in $s'$, and $Q_{\bar{\phi}}$ is a target Q-function that uses old parameters $\bar{\phi}$ from earlier in training, which helps stabilize learning \citep{van2016doubleQlearning}.
These old parameters are periodically updated with the most recent ones $\phi$.
The training algorithm is described in Algorithm~\ref{alg:q_learning}.
Eventually, the Q-value of each candidate response is computed at runtime based on the current state of the conversation. The score given to each candidate response is defined to be its Q-value.
Experiments with this model are described in Section~\ref{seq:exp&eval}.

\subsection{Selecting one response}
\label{subseq:selection}

After scoring each of the candidate messages, the system must pick only one response for the user.
At the time of the \emph{ConvAI} challenge, the feed-forward classifier (described in Section~\ref{subsubseq:sup_scorer}) is used to score candidate messages.
Since its validation accuracy is only $64.23\%$ (when a random binary classifier can achieve $50\%$), a rule-based selection mechanism is built to help the system choose a response.
Other scoring mechanisms explored after the competition are described in Section~\ref{seq:exp&eval}.

\begin{figure}
\centering
    \includegraphics[width=0.7\textwidth]{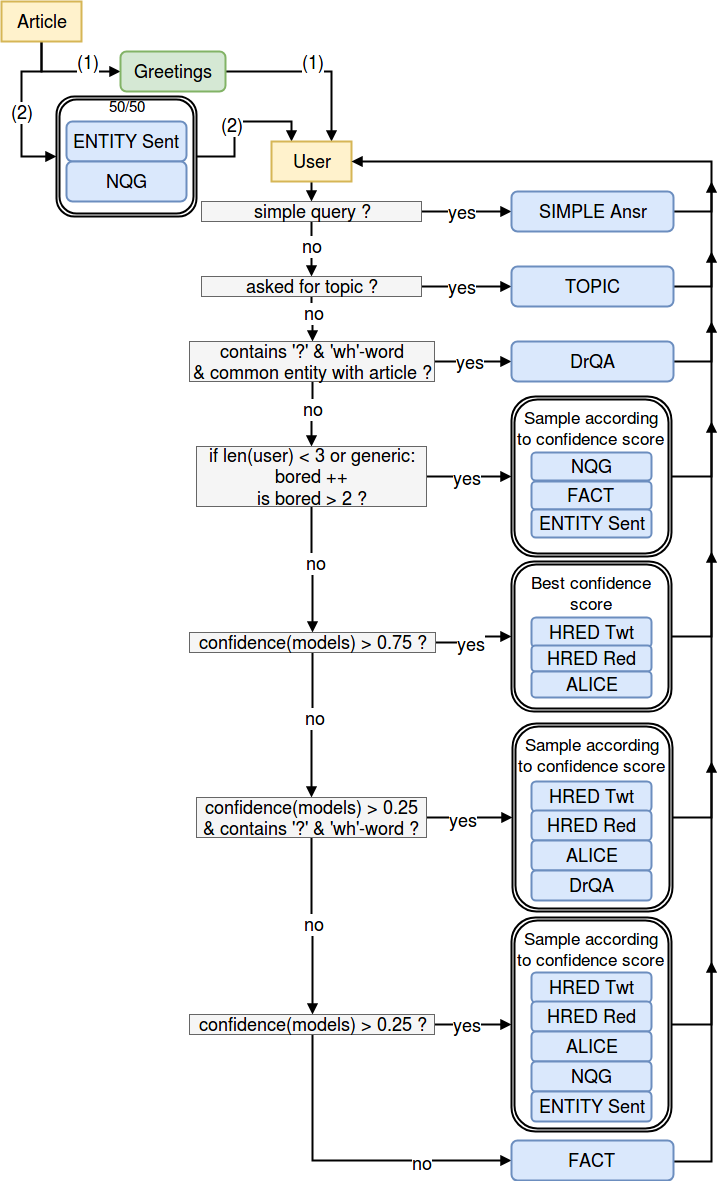}
\caption[Rules followed during a conversation in order to decide which response to return to the user.]{Set of rules to follow during a conversation in order to decide which response to return to the user based on both the conversation state and the score of each possible candidate response. The random article first comes in the conversation, \emph{RLLChatbot} then greets the user and asks a question. After each user's response these rules are followed.}
\label{fig:selection}
\end{figure}

Once a random wiki-news article's paragraph is sent to both users, \emph{RLLChatbot} follows the same pattern: it welcomes the participant with a scripted greeting message (``\textit{Hello! I hope you're doing well. I am doing fantastic today! Let me go through the article real quick and we will start talking about it.}''); and sends a message from either the \emph{Neural Question Generator} (NQG) model (Section~\ref{subsubseq:nqg}) or the \emph{Entity Sentences} model (Section~\ref{subsubseq:entity_sents}).
One of these two models is randomly picked because they are best suited to start a conversation by asking an open question.
The user is then free to answer. Based on the user's reply, \emph{RLLChatbot} generates a set of candidate responses, scores each of them, and returns a message based on the rules described below in order of specificity (also illustrated in Figure~\ref{fig:selection}).
\begin{enumerate}
  \item The most specific module from the ensemble
  is the rule-based \emph{Simple Answers} model (Section~\ref{subsubseq:simple_answ}). The user's message is thus parsed with the set of regular expressions from the model.
  If there is a match, the corresponding pre-defined answer is returned. If there are none, the next rule is applied.
  \item The next specific item that needs to be checked is if the user asks about the topic of the article.
  If the user's last message matches the set of regular expressions designed to catch such messages, a response from the \emph{Topic Classifier} model (Section~\ref{subsubseq:topic}) is returned. If not, the next rule is applied.
  \item Another important and common scenario expected to happen is when the user asks a question about the article.
  It is important to catch those cases because the \emph{DrQA} model (Section~\ref{subsubseq:drqa}) is specifically designed to answer questions.
  As such, if the previous user's message has a common entity with the article, terminates by a question mark, and has a `wh-'word,
  a response from the \emph{DrQA} model (Section~\ref{subsubseq:drqa}) is returned. If the user's response does not match those characteristics, the next rule is applied.
  \item In order to keep the conversation interesting for the participant, a `bored' counter is introduced to see when the user could be bored and remedy that. This counter is incremented every time the user response is short (less than 3 words) or is entirely made of stop-words. As soon as this counter reaches 2,
  a response from the \emph{NQG} model (Section~\ref{subsubseq:nqg}), or the \emph{Fact Retriever} model (Section~\ref{subsubseq:fact_ret}), or the \emph{Entity Sentences} model (Section~\ref{subsubseq:entity_sents}) is sampled according to its score.
  The counter is then reset to 0.
  Only these models from the ensemble are sampled as they are best suited to re-launch the conversation and potentially start talking about something new. If the counter does not reach 2 or if the user is not considered `bored', the next rule is applied.

  
  \item If a candidate response from one of the \emph{HRED} models (Section~\ref{subsubseq:hred}) or the \emph{A.L.I.C.E.} model (Section~\ref{subsubseq:alice}) has a high score (between 0.75 and 1.0), the candidate with the highest score is returned. The motivation is that if the scoring mechanism is strongly confident about a specific response, that response should be returned. These three models are only considered from the ensemble as they are the most flexible and produce generic conversations. If none of these responses have a high score, the next rule is applied.
  \item
  User messages are now split into two categories: they either ask a question, or they do not. If the user asked a question, the same generic models as in the previous rule are considered, with the addition of the \emph{DrQA} model (Section~\ref{subsubseq:drqa}) that is specifically trained to answer questions.
  More formally, if the user message terminates by a question mark and has a `wh-'word, a response from either one of the \emph{HRED} models, or the \emph{A.L.I.C.E.} model, or the \emph{DrQA} model is sampled based on its score
  (as long as it is greater than 0.25). These models are considered because they are all flexible in terms of their type of responses and thus potentially capable of answering a broad range of questions. 
  If the user message does not fall into these characteristics, the next rule is applied.
  \item When the user message does not contain a question, the same models as in the previous rule are sampled except for the \emph{DrQA} model being replaced by the \emph{NQG} model (Section~\ref{subsubseq:nqg}) and the \emph{Entity Sentences} model (Section~\ref{subsubseq:entity_sents}). \emph{DrQA} is not considered because it is designed to answer questions, while the \emph{NQG} and \emph{Entity Sentences} models are designed to ask questions.
  \item Eventually, if none of the above scenario is available (i.e. most responses have a score below 0.25), a more or less related fact from the \emph{Fact Retriever} model (Section~\ref{subsubseq:fact_ret}) is returned. This rule is introduced as a safe `\textit{exit door}' for the system so that it always has something to say.
\end{enumerate}
After selecting a response to return to the user, \emph{RLLChatbot} waits again for the user to reply and selects the next response based on the same set of rules following the same order, until the participant decides to terminate the conversation.

Eventually, the goal is to remove most of the above rules and let the scoring machine completely decide which response to return to the user. 
All these rules can then be replaced by either a \textit{hard} decision: $resp = \arg\max_r Score(r | article, context)$ or a \textit{soft} decision by sampling a response rather than always taking the best one: $resp \sim Score(r | article, context)$. This can make the system slightly more flexible and less reliant on human expertise about how conversations are supposed to go.
To that end, an additional dataset of conversations is collected, as described in Section~\ref{seq:data_collect}, and other scoring mechanisms are considered in Section~\ref{seq:exp&eval}.

%% file: sections/6-data.tex
\section{Data Extension}
\label{seq:data_collect}

This section presents the additional data independently collected after the \emph{ConvAI} challenge in order to improve the scoring and selection process.

\subsection{Collection Procedure}
\label{subseq:collect_procedure}

In order to expand the dataset collected during the \emph{ConvAI} challenge, Facebook's \emph{ParlAI} framework\footnote{\href{http://parl.ai/}{http://parl.ai/}} was used to ask workers from \emph{Amazon Mechanical Turk} to chat with the \emph{RLLChatbot}.
To follow as much as possible the challenge structure, every conversation starts with a random paragraph from a random \emph{SQuAD} article \citep{rajpurkar2016squad}, presents the same greeting message used at the time of the competition, and starts by asking a question about the article's paragraph by running both the \emph{NQG} model (Section~\ref{subsubseq:nqg}) and the \emph{Entity Sentences} model (Section~\ref{subsubseq:entity_sents}).

\begin{figure}
\centering
    \includegraphics[width=1.\textwidth]{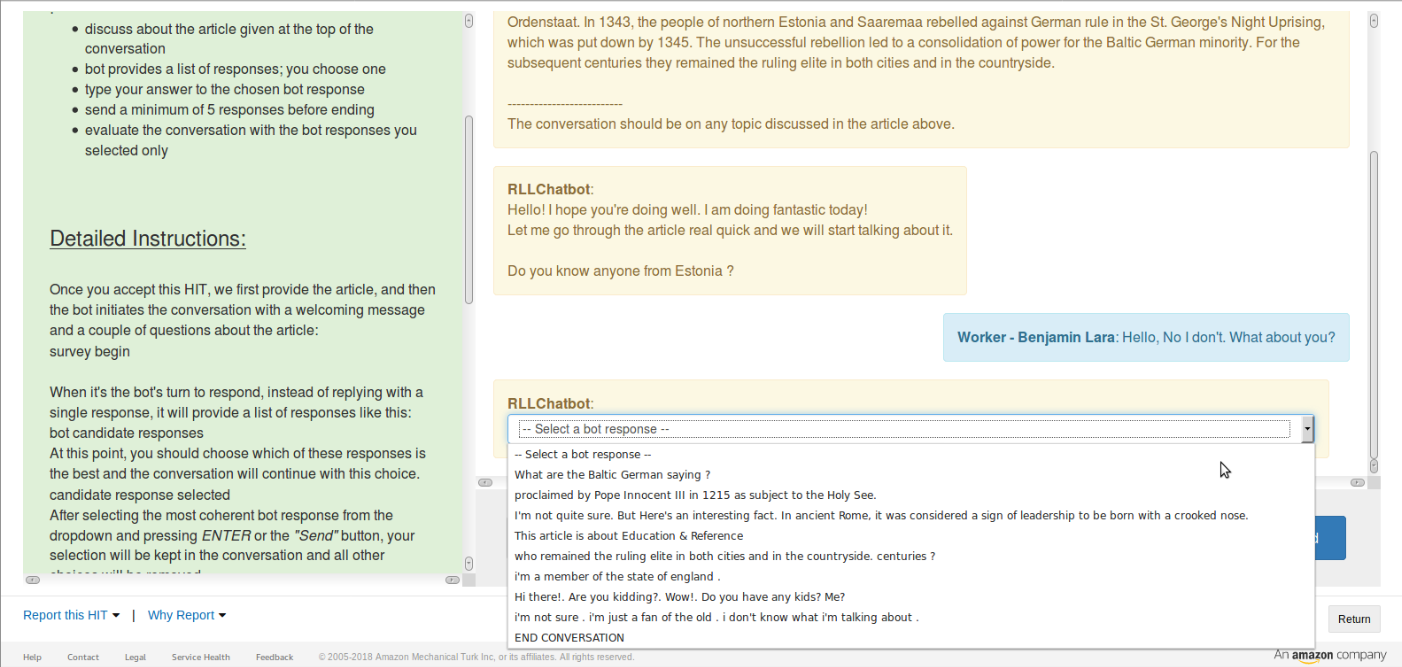}
\caption[User interface of \emph{Amazon Mechanical Turk} during the data collection.]{User interface of \emph{Amazon Mechanical Turk} during the data collection phase. A random paragraph from a random \emph{SQuAD} article and a greeting message are sent. The user then selects the bot response `\textit{Do you know anyone from Estonia?}' and replies. The user is again presented with a choice of potential responses. He has to pick one. Detailed instructions are available on the left of the screen for the user to refer at any time during the conversation.}
\label{fig:amt_interaction}
\end{figure}

The difference with the competition is that in this case, rather than asking the user to up- or down-vote the response that the rule-based selection criteria would have chosen, \emph{the user decides} which candidate response he or she prefers. After selecting \textbf{one} response, the user writes a reply to the chatbot message he or she previously selected. Every following interaction follows the same pattern: the participant is presented a list of candidate responses (all nine models described in Section~\ref{seq:system_overview} are used),
picks one, and writes his or her reply. A visual description of the user interface used during the data collection can be seen in Figure~\ref{fig:amt_interaction}.

After a minimum of 5 interactions, the user can decide to finish the conversation, in which case he is asked to provide a score between 1 (`very bad'), 2 (`bad'), 3 (`medium'), 4 (`good') and 5 (`very good') for the entire conversation. The participant is specifically asked to ignore bot responses that were not selected during the chat when giving this final score. That way, it represents a fair evaluation of the actual conversation that the user just had.

It is important to note that the nine different models that generate candidate responses (described in Section~\ref{seq:system_overview}) were built and trained only once before the competition and remained constant thereafter. This allows to fix the generation intelligence of the system during the entire data collection process and work with a stable environment in which different scoring and selection mechanisms can be compared.
In addition, if a model has many possible candidate responses for each time step in the conversation (such as the \emph{NQG} model), a never before presented candidate response is picked randomly. 

Letting the user decide which response the system returns avoids boring the participant with exploratory selection behaviors from the chatbot, and most importantly, is more data efficient. Indeed, for each interaction, both selected and non-selected responses are saved. On the other hand, during the \emph{ConvAI} challenge, only 1 response per interaction was saved with its corresponding vote.
Here, the \emph{selected} response is considered as \textit{up-voted}, and all the \emph{non-selected} responses as \textit{down-voted}.

\subsection{Data Analysis}
\label{subseq:data_analysis}

\begin{figure}
\centering
    \includegraphics[width=1.\textwidth]{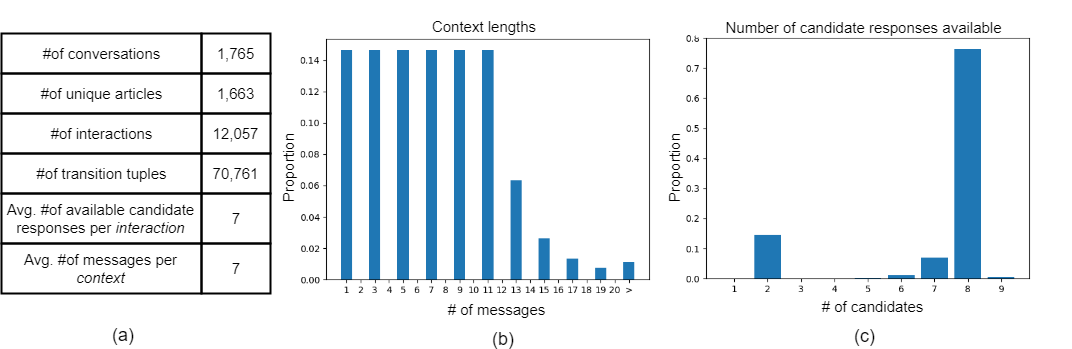}
\caption[Statistics on the data collected with \emph{Amazon Mechanical Turk}.]{Statistics on the entire additional data collected with \emph{Amazon Mechanical Turk}. One \textit{interaction} consists of one bot response followed by a human response. A \textit{transition tuple} is of the form \{article, context, candidate, score\} with the \textit{score} being $1$ (selected) or $0$ (ignored). (a) Table of different data quantities. (b) Proportion of the number of messages per context. (c) Proportion of the number of available candidate responses per interaction.}
\label{fig:amt_data_stats}
\end{figure}

The statistics of the data collected can be seen in Figure~\ref{fig:amt_data_stats}.
Figure~\ref{fig:amt_data_stats}(b) shows that there is an equal number of data instances with context length being 1, 3, 5, 7, 9, and 11. That is to be expected since participants are asked to perform a minimum of 5 interactions before ending the conversation.
It can also be seen that a few users continued their conversation further, which is a sign of appreciation for the \emph{RLLChatbot}. 
The average number of interactions per chat is \texttildelow 7 from Table~\ref{fig:amt_data_stats}(a).
Moreover, there is only an odd number of messages in every context because the first greeting message from the chatbot is the first context message; then the user picks a candidate message and replies to it, thus adding two messages to the context. Thereafter every one interaction is made of one bot message followed by one user reply.

Figure~\ref{fig:amt_data_stats}(c) shows that most of the time (76.32\% of the data instances) the number of candidate responses available to the user is 8. That is to be expected since from the nine generative models, the \emph{Simple Answers} model (Section~\ref{subsubseq:simple_answ}) provides a response only when the previous user message matches a small set of regular expressions. 
All the other models are expected to return a response at all times.
Note that in 14.64\% instances, only two candidate responses are shown to the user, which is also to be expected since at every beginning of conversation only the \emph{NQG} and the \emph{Entity Sentences} models are triggered as previously explained.
However, some rare times the number of candidate responses available to the user is only 7, 6, or even 5 (8.48\% of the data instances in total). This is due to some models not providing responses.

\begin{figure}
\centering
    \includegraphics[width=1.\textwidth]{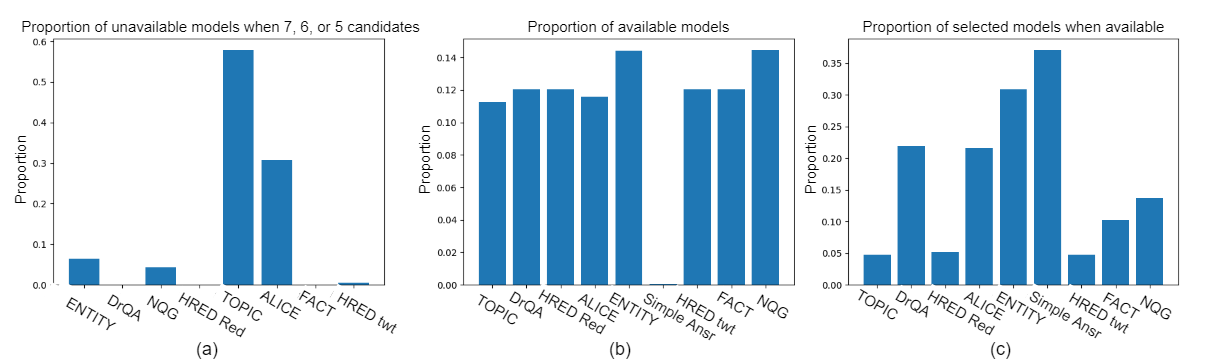}
\caption[Statistics on the different candidate models usage.]{Statistics on the different candidate models usage. (a) Proportion of the number of times each model (except the \emph{Simple Answers} model) is \emph{not} available when the number of candidate responses is 5, 6, and 7. (b) Proportion of the number of times each model is in the list of available candidate responses. (c) Proportion of the number of times each model is selected \emph{when it is in the list of available candidate responses}.}
\label{fig:amt_data_models}
\end{figure}

To understand a little more why this happens, Figure~\ref{fig:amt_data_models}(a) presents statistics about which model is not available to the user when there are only 5, 6, or 7 candidate responses (ignoring the \emph{Simple Answers} model that is only available in rare cases). The \emph{Topic Classifier} (Section~\ref{subsubseq:topic}) and the \emph{ALICE} model (Section~\ref{subsubseq:alice}) are mostly responsible for cases in which the user is missing some candidate responses (\texttildelow 88.62\% of the time).

Some investigation revealed that the \emph{fastText} topic classifier \citep{joulin2016fasttext} used is running in a separate thread from the actual \emph{Topic Classifier} module.
Thus, if the \emph{fastText} classifier thread is delayed by other modules, it may not have a label for the article's paragraph, while the \emph{Topic Classifier} module is ready to propose a candidate response. In such cases, the \emph{Topic Classifier} model does not wait for the \emph{fastText} classifier to provide a topic and returns an empty message instead.

The \emph{ALICE} model is the second model that most often fails to return a response. After investigating the original source code downloaded for this model, we noticed that when \emph{ALICE} initially generates a response containing quotation marks, it assumes that the response has profanity language and returns an empty message instead. Some failure examples are shown in Table~\ref{table:alice_fails}.
Other models that rarely fail to return an answer are the \emph{Entity Sentences} (Section~\ref{subsubseq:entity_sents}), and the \emph{NQG} (Section~\ref{subsubseq:nqg}) models. The most common reason for failure in those cases is when the article's paragraph is one sentence. Indeed, in short paragraphs there is not always enough information for these models to return a response.

Furthermore, Figure~\ref{fig:amt_data_models}(b) of shows some statistics about which models are available to the user for all interactions. As expected, the \emph{Entity Sentences} and \emph{NQG} models are more often present since they are the only ones presented at the first turn.
As mentioned above, the \emph{Topic Classifier} model and the \emph{ALICE} model are slightly less presented than the other candidate models.
In addition, the \emph{Simple Answers} model is almost never presented to the user since the set of regular expressions used to fire this model is short and not flexible.

\begin{table}
\centering
\begin{tabular}{|l|l|}
\cline{1-2}
  \multicolumn{1}{|c|}{\textbf{User message}} & \multicolumn{1}{|c|}{\textbf{\textit{ALICE} possible responses}} \\ \cline{1-2}
  Two & \begin{tabular}[c]{@{}l@{}}Bertrand Russell once said that ``it must \\ have required many ages to discover \\ that a brace of pheasants and a couple of\\ days were both instances of the number two''.\end{tabular} \\ \cline{1-2}
  \begin{tabular}[c]{@{}l@{}}I think it would be nice, but I don't \\ know a whole lot about it \end{tabular} & Why do you want nice?. ``it'' being nice? \\ \cline{1-2}
  We don't follow prodigies anymore & By ``we'' do you mean you and me? \\ \cline{1-2}
  I have no idea & ``?'' No idea about that? \\ \cline{1-2}
  \begin{tabular}[c]{@{}l@{}}Yes in a way. But I think it's \\ more an historical article. \end{tabular} & \begin{tabular}[c]{@{}l@{}}Great.. In what way? What does ``it'' refer to?\\ \\ Next question?. In what way? It depends on \\ what the meaning of the word ``it'' is.\\ \\ Aha.. In what way? ``It'' being what?\end{tabular} \\ \cline{1-2}
\end{tabular}
\caption{Some user messages that fails the \emph{ALICE} model: since the possible responses on the right column contain quotation marks, \emph{ALICE} ignored those valid responses and returned an empty message instead.}
\label{table:alice_fails}
\end{table}

Eventually, it is important to understand which model is the most often chosen by human users talking to the \emph{RLLChatbot}. To that end, Figure~\ref{fig:amt_data_models}(c) presents the proportion of the number of times each model is selected \emph{when it actually appears in the list of suggested candidate responses}.
The model with the highest selection rate is the \emph{Simple Answers} model with a score of 37\%. This is reassuring in a sense because even though this model is rarely presented to the user, when it is, its manually defined responses are preferred by users.
The next most chosen model is the \emph{Entity Sentences} model with a selection rate of 30.81\% when available.
The next two models with comparable selection rates are \emph{DrQA} and \emph{ALICE} with 21.87\% and 21.61\% selection rates respectively. The three rule-based systems of the ensemble are in this top-four. It is thus clear that, in general, the preferred systems according to human users are rule-based systems.
The remaining models in decreasing order are \emph{NQG} with 13.69\% selection rate, \emph{Fact Retriever} with 10.19\%, and \emph{HRED} reddit, \emph{HRED} twitter, and \emph{Topic Classifier} with 5.21\%, 4.77\% and 4.71\% selection rates respectively.
This analysis shows that current fully generative sequence-to-sequence models have not yet reached the user preference level of more restrictive, rule-based, or even retrieval-based models.

%% file: sections/7-experiments.tex
\section{Experimentation and Evaluation}
\label{seq:exp&eval}

After describing and analyzing in detail the dataset collected from \emph{Amazon Mechanical Turk}, this section presents various experiments on this data and reports results on a held-out test set.

\subsection{Experiments}
\label{subseq:experiments}

The reported experiments aim at building a mechanism that can automatically select which response to return to the user from a set of previously generated candidate responses.
In Section~\ref{seq:scorer} two different neural network architectures are presented: one using hand-crafted features (Figure~\ref{fig:mlp_scorer}), the other using Gated Recurrent Unit networks to automatically extract features (Figure~\ref{fig:q_scorer}).
Two different training algorithms are also described: one using the cross-entropy classification loss (Equation~\ref{eq:cross_entropy}), the other using the Huber loss with fitted Q-iteration (Algorithm~\ref{alg:q_learning}).
Eventually, three selection criteria are mentioned: one rule-based process (Figure~\ref{fig:selection}), and two heuristics based on either taking the response with maximum score or sampling one according to its score.
All these different ideas are now combined in the following set of experiments:
\begin{itemize}
  \item \textbf{SmallR}: The feed-forward network with hand-crafted features is trained to predict the immediate reward of a given candidate response: either 0 or 1. Equation~\ref{eq:cross_entropy} is used.
  \item \textbf{DeepR}: The Gated Recurrent Unit network (GRU) is trained to predict the immediate reward of a given candidate response: either 0 or 1. Equation~\ref{eq:cross_entropy} is used, but with the GRU architecture (Figure~\ref{fig:q_scorer}).
  \item \textbf{SmallQ}: The feed-forward network with hand-crafted features is trained to predict the Q-value of a given candidate response. Algorithm~\ref{alg:q_learning} is used, but with the architecture described in Figure~\ref{fig:mlp_scorer}.
  \item \textbf{DeepQ}: The Gated Recurrent Unit network is trained to predict the Q-value of a given candidate response. Algorithm~\ref{alg:q_learning} is used.
\end{itemize}
All these experiments yield a scoring model that is able to score candidate responses. On top of these, three different selection mechanisms are explored:
\begin{itemize}
  \item \textbf{Rule-Based}: The hand-crafted rules as described in Section~\ref{subseq:selection} and in Figure~\ref{fig:selection} decide which response is selected.
  \item \textbf{Sampled}: A random candidate response is sampled (without replacement\footnote{k responses are sampled when evaluating the model with \textit{Recall@k}.}) according to the distribution given by the scores of all candidate responses.
  \item \textbf{Argmax}: The candidate response with the highest score is selected (without replacement).
\end{itemize}

The data described in Section~\ref{seq:data_collect} is split into 80\% training, 10\% validation, and 10\% testing set.
Both the validation and the testing set have no overlapping news article with the training set. This gives $1,330$ unique articles in the training set, $165$ in the validation set, and $168$ in the testing set. Further details can be found in Table~\ref{table:amt_data}.
For instance, from the total $70,761$ examples, only $10,292$ have positive reward (+1), while $60,469$ have negative reward (0). This is due to the fact that, for each interaction, both the uniquely selected candidate response (labeled as positive example) and all other non-selected candidate responses (labeled as negative examples) are collected.
Therefore a second version of the training set is constructed by over-sampling positive examples as one can see in the third column of Table~\ref{table:amt_data}.
The over-sampled training set is used in all experiments regarding the classification of candidate responses (\emph{SmallR} and \emph{DeepR} experiments). Both the over-sampled and the regular training sets are experimented for the estimation of Q-values (\emph{SmallQ} and \emph{DeepQ} experiments). For all experiments, a random search of 100 parameter combination is done. Details about the explored parameters can be found in Appendix~\ref{subapp:param_explored}.

\begin{table}
\centering
\begin{tabular}{|l|l|l|l|l|l|}
\cline{1-6}
  \multicolumn{1}{|c|}{\textbf{}} & \begin{tabular}[c]{@{}l@{}}\textbf{Entire}\\ \textbf{data}\end{tabular} & \begin{tabular}[c]{@{}l@{}}\textbf{Training set}\\ \textbf{(regular)}\end{tabular} & \begin{tabular}[c]{@{}l@{}}\textbf{Training set}\\ \textbf{(over-sampled)}\end{tabular} & \begin{tabular}[c]{@{}l@{}}\textbf{Validation}\\ \textbf{set}\end{tabular} & \begin{tabular}[c]{@{}l@{}}\textbf{Testing}\\ \textbf{set}\end{tabular} \\ \cline{1-6}
  \textbf{unique articles} & 1,663 & 1,330 & 1,330 & 165 & 168 \\ \cline{1-6}
  \textbf{all examples} & 70,761 & 56,564 & 96,659 & 7,114 & 7,083 \\ \cline{1-6}
  \textbf{positive examples} & 10,292 & 8,233 & 48,328 & 1,031 & 1,028 \\ \cline{1-6}
  \textbf{negative examples} & 60,469 & 48,331 & 48,331 & 6,083 & 6,055 \\ \cline{1-6}
\end{tabular}
\caption{Statistics on the regular training set, over-sampled training set, validation set and testing set.}
\label{table:amt_data}
\end{table}

The best parameter combination for the reward classification scorers (\emph{SmallR} and \emph{DeepR} experiments) is searched\footnote{Further details about parameter exploration can be found in Appendix~\ref{subapp:param_explored}.} by evaluating the F1 score on the validation set with early stopping and a patience of 20 epochs.
Training is stopped based on the validation F1 score rather than the validation accuracy because of the imbalance in the validation set as one can see in the fourth column of Table~\ref{table:amt_data}.
After running 100 \emph{SmallR} experiments and another 100 \emph{DeepR} experiments, the best combination of parameters gave a validation F1 score of $0.420$ for the best \emph{SmallR} model, and a validation F1 score of $0.373$ for the best \emph{DeepR} model. The different parameters yielding these results can be found in Appendix~\ref{subapp:best_r_params}.


The best parameter combination for the Q-value estimation scorers (\emph{SmallQ} and \emph{DeepQ} experiments) is searched\footnote{Further details about parameter exploration can be found in Appendix~\ref{subapp:param_explored}.} by evaluating the Huber loss on the validation set with early stopping and a patience of 20 epochs.
After running 100 \emph{SmallQ} experiments and another 100 \emph{DeepQ} experiments, the best combination of parameters gave a minimal validation loss of $0.00488$ for the best \emph{SmallQ} model, and a minimal validation loss of $0.00505$ for the best \emph{DeepQ} model. The different parameters yielding these results can be found in Appendix~\ref{subapp:best_q_params}.

\subsection{Evaluation}
\label{subseq:eval}

To automatically evaluate how the above models perform, the conversational dataset collected in Section~\ref{seq:data_collect} is used to measure how well each model can predict which response was chosen by the human.
The held-out test set is used for evaluation.
Similarly to the \emph{Next Utterance Classification} task \citep{lowe2016NUC}, the \textit{Recall@k} (\textit{R@k}) is measured, which is the success rate of finding the correct response in the top k responses ranked in order according to the scoring model.
All the above experiments (\emph{SmallR}, \emph{DeepR}, \emph{SmallQ}, \emph{DeepQ}) as well as the initial baseline model are evaluated.
Three different selection mechanisms are considered: \emph{Rule-Based}, \emph{Argmax}, and \emph{Sampled}.
Results can be seen and compared in Table~\ref{table:detailed_results}. The following sections discuss in details these results.

\begin{table}
\centering
\begin{tabular}{|l|l|l|l|l|l|l|l|}
\cline{1-8}
 & \textbf{\begin{tabular}[c]{@{}l@{}}R@1\\ rule based\end{tabular}} & \textbf{\begin{tabular}[c]{@{}l@{}}R@1\\ argmax\end{tabular}} & \textbf{\begin{tabular}[c]{@{}l@{}}R@1\\ sampled\end{tabular}} &  & \textbf{\begin{tabular}[c]{@{}l@{}}Avg. R@k\\ rule based\end{tabular}} & \textbf{\begin{tabular}[c]{@{}l@{}}Avg. R@k\\ argmax\end{tabular}} & \textbf{\begin{tabular}[c]{@{}l@{}}Avg. R@k\\ sampled\end{tabular}} \\ \cline{1-8}
\textbf{Baseline} & \textbf{28.89} \% & 20.33 \% & 19.94 \% &  & 73.53 \% & \textbf{73.62} \% & 70.40 \% \\ \cline{1-8}
\textbf{SmallR} & 36.87 \% & \textbf{39.20 \%} & 22.76 \% &  & 78.60 \% & \textbf{82.44} \% & 71.03 \% \\ \cdashline{1-4}\cdashline{6-8}
\textbf{DeepR} & 37.16 \% & 37.26 \% & 20.91 \% &  & 77.88 \% & 80.78 \% & 69.78 \% \\ \cline{1-8}
\textbf{SmallQ} & 24.32 \% & 16.73 \% & 19.16 \% &  & 70.50 \% & 63.64 \% & 66.06 \% \\ \cdashline{1-4}\cdashline{6-8}
\textbf{DeepQ} & \textbf{24.61} \% & 16.63 \% & 21.01 \% &  & \textbf{70.54} \% & 64.07 \% & 66.86 \% \\ \cline{1-8}
\end{tabular}
\caption{\emph{Recall@1} and average \emph{Recall@k} for all experiments and all selection mechanisms.}
\label{table:detailed_results}
\end{table}

\subsubsection{Baseline}

\begin{figure}
\centering
    \includegraphics[width=1.0\textwidth]{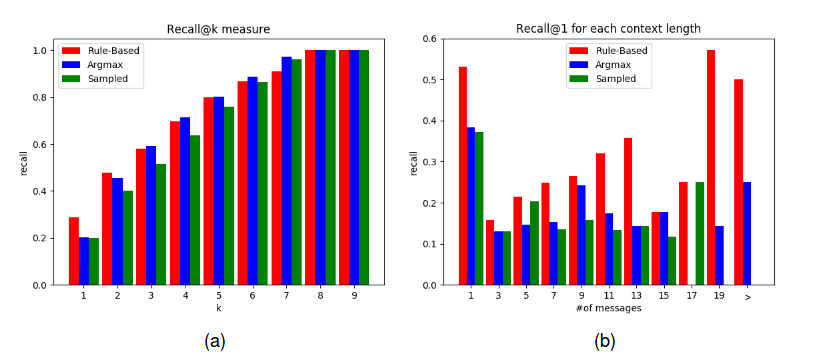}
\caption[Recall measurements for different selection mechanisms with the baseline model.]{Recall measurements for different selection mechanisms with the baseline model used during the \emph{ConvAI} challenge. (a) \textit{Recall@k} for all possible values of k. (b) \textit{Recall@1} for different context lengths. The number of candidate responses vary between 2 and 9 as described in Figure~\ref{fig:amt_data_stats}(c).}
\label{fig:baseline_recalls}
\end{figure}

The baseline is defined to be the scoring network described in Section~\ref{subsubseq:sup_scorer} and trained for the \emph{ConvAI} challenge with a validation accuracy of $64\%$ on the \emph{ConvAI} competition dataset described in Section~\ref{subseq:competition_data}.
Measurements of the \textit{Recall@k} metric can be seen in Figure~\ref{fig:baseline_recalls}(a). Hand-crafted rules described in Section~\ref{subseq:selection} helped a lot to boost the \textit{Recall@1} score for this model jumping from $20\%$ with \emph{Argmax} selection to $29\%$ with \emph{Rule-Based} selection. This is a good sign for the hand-crafted rules as it shows that the external knowledge used when designing them is useful for selecting appropriate responses.
It is also interesting to see that starting at \textit{R@3}, \emph{Argmax} selection becomes better than \emph{Rule-Based} selection.


Figure~\ref{fig:baseline_recalls}(b) focuses on the \textit{R@1} score, but for different context lengths, that is, at different time steps in a conversation.
As expected, the \emph{Rule-Based} selection mechanism yields better scores than the other two at all times in a conversation.
Recall that for contexts of length 1, the number of candidate responses is only two, thus the random selection process of the \emph{Rule-Based} systems (see Section~\ref{subseq:selection}) is expected to give a score around $50\%$.
For contexts of length greater than 1 (the number of candidate responses is around 8), the recall score increases as the conversation goes on, up until 13 messages in the context. This is specifically true for the \emph{Rule-Based} selection mechanism, reaching a \textit{R@1} score of $36\%$ in conversations with 13 messages in the context.
This may be due to the fact that with longer conversations, the system has more information about the nature of the conversation and is thus better suited to select which response is the most appropriate.
Longer contexts (more than 13 messages) are not frequent enough in the dataset (as one can see from Figure~\ref{fig:amt_data_stats}(b)) to have a meaningful interpretation about the \textit{R@1} score.

\subsubsection{Classifiers}

\begin{figure}
\centering
    \includegraphics[width=1.0\textwidth]{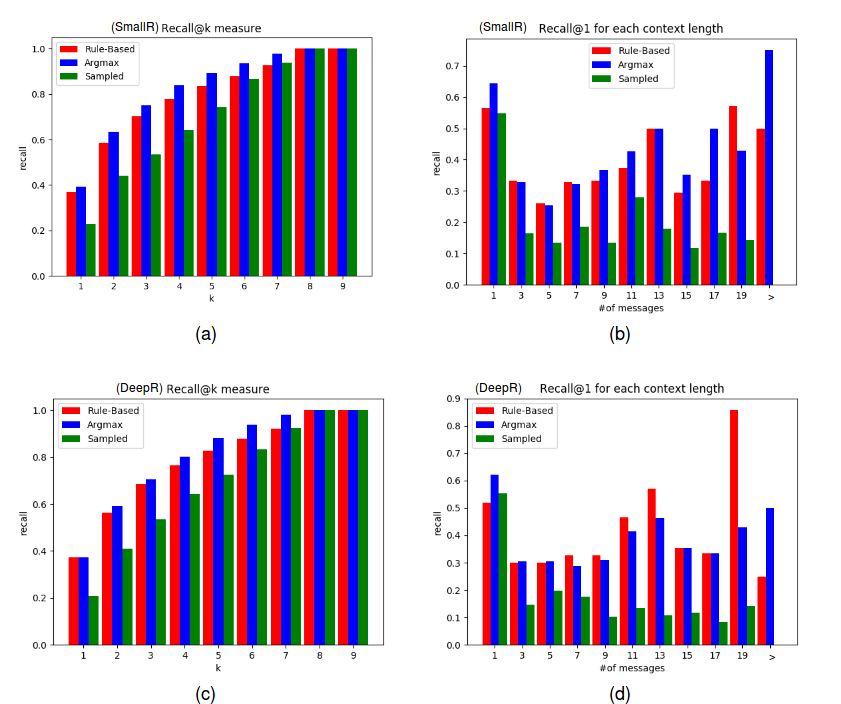}
\caption[Recall measurements for different selection mechanisms with the best \emph{SmallR} and \emph{DeepR} models.]{Recall measurements for different selection mechanisms with the best \emph{SmallR} and \emph{DeepR} models. (a) \textit{Recall@k} for all possible values of k with the best \emph{SmallR} model. (b) \textit{Recall@1} for different context lengths with the best \emph{SmallR} model. (c) \textit{Recall@k} for all possible values of k with the best \emph{DeepR} model. (d) \textit{Recall@1} for different context lengths with the best \emph{DeepR} model. The number of candidate responses vary between 2 and 9 as described in Figure~\ref{fig:amt_data_stats}(c).}
\label{fig:R_recalls}
\end{figure}

Figure~\ref{fig:R_recalls} presents different recall measurements for the best \emph{SmallR} and \emph{DeepR} models with a maximal validation F1 score of $0.420$ and $0.373$ respectively.
The first thing to notice is the improvement with the baseline model presented earlier from a $29\%$ \textit{R@1} score to a $39\%$ \textit{R@1} score for the best \emph{SmallR} model, and to a $37\%$ \textit{R@1} score for the best \emph{DeepR} model.
In addition, Figures~\ref{fig:R_recalls}(a) and \ref{fig:R_recalls}(c) show that the \emph{Argmax} selection mechanism is as good if not better than the custom \emph{Rule-Based} selection mechanism for all values of $k$ in \textit{R@k} scores.
This means that the system is now much more flexible as it does not rely on human rules to select which message to return to the user.
Figures~\ref{fig:R_recalls}(a) and \ref{fig:R_recalls}(c) also show that the best \emph{SmallR} model and the best \emph{DeepR} model are similar in terms of \textit{Recall@k} score as it grows at the same rate as $k$ increases.
However, the \emph{SmallR} model is slightly better, specially with the \emph{Argmax} selection mechanism which has an average recall score (computed by taking the average \textit{R@k} score over all values of $k$) of $82.44\%$ against $80.78\%$ for the best \emph{DeepR} model with \emph{Argmax} selection as Table~\ref{table:detailed_results} reports.
This shows that the deeper architecture involving GRU networks captures meaningful information about the state of the conversation, but the hand-crafted features are still slightly better in those experiments.
Furthermore, the deeper architecture being designed more specifically for predicting Q-values by decomposing state and action values, one can expect that this complication may not be optimal for the current classification task.

Eventually, Figures~\ref{fig:R_recalls}(b) and \ref{fig:R_recalls}(d) report the \textit{R@1} score of both models with different selection mechanisms at different time steps in the conversation.
One can see that the main advantage of the \emph{Argmax} selection over the \emph{Rule-Based} selection reported from Figures~\ref{fig:R_recalls}(a) and \ref{fig:R_recalls}(c) actually happens at the beginning of the conversation when the context length is 1.
When the discussion contains more messages, the \emph{Rule-Based} selection mechanism is sometimes better, sometimes worst than the \emph{Argmax} selection.
In general though, as in the baseline model, the \textit{R@1} score tends to increase with context length, up until 13 messages.
Longer conversations are not frequent enough in the dataset (as one can see from Figure~\ref{fig:amt_data_stats}(b)) to have a meaningful \textit{R@1} score interpretation.

\begin{figure}
\centering
    \includegraphics[width=1.0\textwidth]{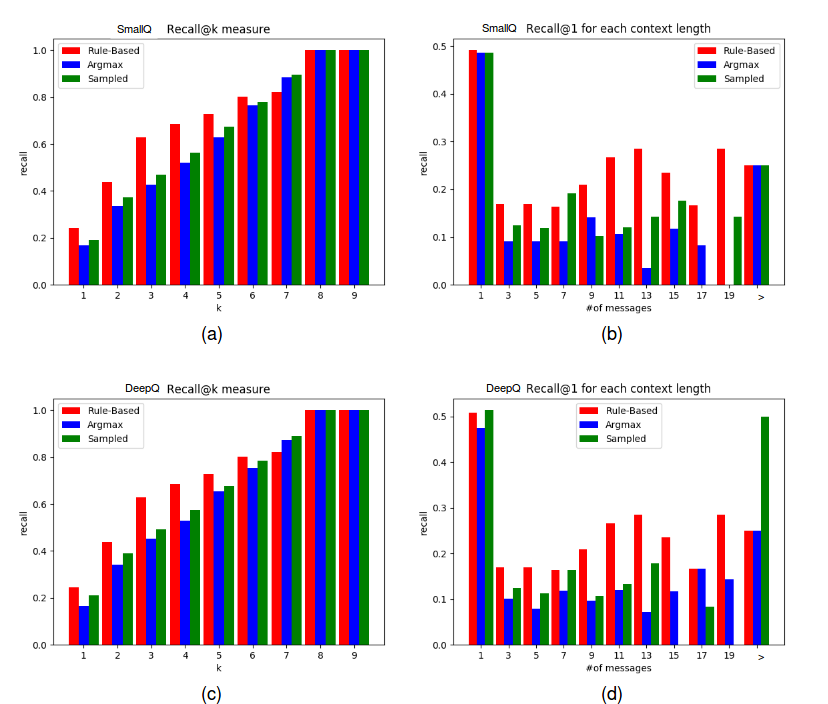}
\caption[Recall measurements for different selection mechanisms with the best \emph{SmallQ} and \emph{DeepQ} models.]{Recall measurements for different selection mechanisms with the best \emph{SmallQ} and \emph{DeepQ} models. (a) \textit{Recall@k} for all possible values of k with the best \emph{SmallQ} model. (b) \textit{Recall@1} for different context lengths with the best \emph{SmallQ} model. (c) \textit{Recall@k} for all possible values of k with the best \emph{DeepQ} model. (d) \textit{Recall@1} for different context lengths with the best \emph{DeepQ} model. The number of candidate responses vary between 2 and 9 as described in Figure~\ref{fig:amt_data_stats}(c).}
\label{fig:Q_recalls}
\end{figure}

\subsubsection{Q-value Predictors}

Finally, Figure~\ref{fig:Q_recalls} presents different recall measurements for the best \emph{SmallQ} and \emph{DeepQ} models with a minimal validation loss of $0.00488$ and $0.00505$ respectively.
The first thing one can notice is that these models are actually worse than the baseline classifier with a \textit{R@1} score of $25\%$ attained by the best \emph{DeepQ} model, against the $29\%$ baseline score.
The scoring mechanism being poor at evaluating candidate responses, Figures~\ref{fig:Q_recalls}(a) and \ref{fig:Q_recalls}(c) show that the \emph{Rule-Based} selection process is now stronger than the other two for all values of $k$ under 7. 
Overall, the average Recall score of the \emph{SmallQ} model is slightly lower than its counterpart \emph{DeepQ}, specially with the \emph{Argmax} selection mechanism which has an average Recall score (computed by taking the average \textit{R@k} score over all values of $k$) of $63.64\%$ against $64.07\%$ for the best \emph{DeepQ} model with \emph{Rule-Based} selection as Table~\ref{table:detailed_results} reports.
This shows that the deeper architecture involving GRU networks is preferred to estimate Q-values. This is to be expected as the architecture was inspired by \emph{Dueling Deep Q-Networks} \citep{wang2015duelingQDN}.

Another interesting result observed in Figure~\ref{fig:Q_recalls} is that, unlike in the classifier models, the \emph{Sampled} selection process seems to perform better than the \emph{Argmax} selection in both \emph{SmallQ} and \emph{DeepQ} models. This shows that being greedy with respect to the predicted Q-value may not always be the best strategy, and allowing some stochasticity can be beneficial in those cases.
This is another sign that the predicted Q-values are not informative enough to make a greedy decision.
Eventually, Figures~\ref{fig:Q_recalls}(b) and \ref{fig:Q_recalls}(d) indicate that, just like in the previous experiments, longer contexts allow better \textit{R@1} score with the \emph{Rule-Based} selection mechanism (up until 13 messages).

\vskip0.5cm

Overall, the experiments conducted and described above show that the additional data manually collected is indeed informative about how humans pick their responses. This novel dataset can be used to train message scoring and message selection models, thus minimizing the need of human expertise by automatically extracting text features.
In addition, those experiments show that the choice of the scoring algorithm used is critical, as the \emph{SmallQ} and \emph{DeepQ} experiments yield poor results. This can be caused by the fact that the state and action space possible in a conversational environment is enormous if not infinite, and that the collected data cannot possibly capture all of it.

\subsubsection{Qualitative Evaluation}

\begin{table}
\begin{tabular}{|r|l|l|}
  \cline{1-3}
  article & \begin{tabular}[c]{@{}l@{}} \textless{}...\textgreater{}. Dou Wu and the Grand Tutor \\ Chen Fan attempted a coup d'etat \\ against the eunuchs Hou Lan, Cao Jie, \\ and Wang Fu. \textless{}...\textgreater{} General Zhang \\ Huan favored the eunuchs. \textless{}...\textgreater{} \end{tabular} & \begin{tabular}[c]{@{}l@{}} \textless{}...\textgreater{}. Most of these companies \\ \textless{}...\textgreater{} temporarily disable the \\ 10NES chip in the NES. \textless{}...\textgreater{} \end{tabular} \\ \cdashline{1-3}
  bot & Do you know Wu in person? & \begin{tabular}[c]{@{}l@{}} Do you think you would like \\ to work for NES? \end{tabular} \\ \cdashline{1-3}
  user & No, I do not. & Yes I would. \\ \cdashline{1-3}
  bot & \begin{tabular}[c]{@{}l@{}} Did you know that a Boeing 747s \\ wingspan is longer than the Wright \\ brother's first flight.? \end{tabular} & \begin{tabular}[c]{@{}l@{}} The article is related to Computer \\ \& Internet. \end{tabular} \\ \cdashline{1-3}
  user & I didn't know that. & That is interesting. \\ \cdashline{1-3}
  bot & \begin{tabular}[c]{@{}l@{}} Do you know what Chen Fan \\ did in his life? \end{tabular} & I'm a fan of the game. \\ \cdashline{1-3}
  user & \begin{tabular}[c]{@{}l@{}} He was arrested for an \\ attempted coup d'etat. \end{tabular} & What game? \\ \cdashline{1-3}
  bot & \begin{tabular}[c]{@{}l@{}} I don't have a lot of other \\ questions about this article. \\ Maybe you can ask me one? \end{tabular} & \begin{tabular}[c]{@{}l@{}} We are playing Turing's \\ imitation game. \end{tabular} \\ \cdashline{1-3}
  user & Who favoured the eunuchs? & How does it go? \\ \cdashline{1-3}
  bot & General Zhang Huan & How should I know how it go does. \\ \cline{1-3}
  article & \begin{tabular}[c]{@{}l@{}} \textless{}... \textgreater{} Margaret Thatcher travelled \\ to Beijing to negotiate with the \\ Chinese government \textless{}...\textgreater{} \end{tabular} & \begin{tabular}[c]{@{}l@{}} \textless{}...\textgreater{}. The Los Angeles Avengers \\ announced that they were \\ folding the franchise. \textless{}...\textgreater{} \end{tabular} \\ \cdashline{1-3}
  bot & Are you a Chinese ? & \begin{tabular}[c]{@{}l@{}} Have you ever used any of the \\ Los Angeles Avengers's \\ products or services? \end{tabular} \\ \cdashline{1-3}
  user & I am not chinese I'm american & No i have not \\ \cdashline{1-3}
  bot & i like the way you think . i like it . & \begin{tabular}[c]{@{}l@{}} Are you kidding? \\ Do you think you will? \end{tabular} \\ \cdashline{1-3}
  user & That's just my brain & Is it any good? \\ \cdashline{1-3}
  bot & \begin{tabular}[c]{@{}l@{}} Do you know that In ancient \\ Greece idiot meant a \\ private citizen or layman.? \end{tabular} & This article is about Sports \\ \cdashline{1-3}
  user & Ya those silly philosophers & Which one? \\ \cdashline{1-3}
  bot & \begin{tabular}[c]{@{}l@{}} I'm a Romanian and I'm not \\ even a king . \end{tabular} & The article is related to Sports \\ \cdashline{1-3}
  user & \begin{tabular}[c]{@{}l@{}} Well there aren't very many \\ kings nowadays \end{tabular} & Basketball? \\ \cdashline{1-3}
  bot & I'm a king . & I think it's about Sports \\ \cline{1-3}
\end{tabular}
\caption[Examples of four bot-to-human conversations with the final \emph{RLLChatbot}]{Examples of four bot-to-human conversations with the final \emph{RLLChatbot}, using the best \emph{SmallR} scorer and the \emph{Argmax} selection mechanism.}
\label{table:convo_examples}
\end{table}

The best scoring and selection mechanisms are combined together according to all the previously described experiments. The final version of the \emph{RLLChatbot} is using the \emph{SmallR} scorer and the \emph{Argmax} selection with an average \emph{R@k} score of 82.44\%. This version of the system was used to collect a few conversations with human users.
We present in Table~\ref{table:convo_examples} four such conversations.

The top left conversation follows a question -- answer structure where the \emph{RLLChatbot} asks most of the questions with the \emph{Entity Sentences} model. At the end of the conversation the bot is then answering user questions using the \emph{DrQA} model.
The top right conversation is more of a social chat and the topic of the conversation diverges a little from the article but stays coherent.
These first two conversations are some good examples of coherent interactions we collected.

The lower left chat is an example in which the \emph{RLLChatbot} changes topic with the \emph{Fact Retriever} model. In addition, it contradicts itself in the last interaction with the ``I'm a king.'' reply.
The lower right conversation is an example in which the system goes in circles and no progress is made in the chat.
These two conversations are examples of incoherent interactions collected with the dialog agent.

One weakness of using a probabilistic selection mechanism (\emph{Argmax}) is that we cannot explicitly check for contradictions or repetitions from the system, whereas a simple rule based system can avoid those issues. On the other hand, using probabilistic models allow the system to be more flexible to new conversation topics.

%% file: sections/8-conclusion.tex
\section{Conclusion}
\label{seq:conclusion}

\subsection{Summary}
\label{subseq:summary}

Throughout this article we presented the \emph{RLLChatbot}: a conversational agent capable of discussing random news paragraphs with a human user. Seeing a real system in details provide a lot of value to dialog researchers and practitioners.
Using an ensemble of rule-based and statistical models, the system differentiates itself from previous conversational agents in many ways. Being non-goal oriented, it has to be flexible enough to discuss a wide range of topics, which motivated the use of different models ranging in their specificity.

Several models are used to generate up to nine distinct candidate responses at each interaction of a conversation.
The final message returned to the user is selected according to a trained scoring mechanism.
In contrast, typical conversational agents use at least 3 modules to produce a response: a natural language understanding machine, a dialog manager (often made of many sub-modules), and a natural language generator \citep{raux2005prefermodularDS,callejas2005prefermodularDS}.

Another focus of this work is the presentation of a novel conversational dataset collected to train different message scoring mechanisms. Multiple bot responses are available in each interaction of a conversation, and the goal of the machine is to identify which response was chosen by the human. Four initial strategies are presented: two 
relying on supervised learning to perform a classification task,
and two relying on reinforcement learning to perform a prediction task.
Two types of architectures are also considered: one using hand-crafted features with a feed-forward neural network, the other using automatic feature extraction with Gated Recurrent Unit (GRU) networks.
The difference between the deep GRU network and the feed-forward network with hand-crafted features is negligible. 
However, the training algorithm is a critical decision as the models trained to predict Q-values with reinforcement learning techniques are not as powerful as the baseline model according to the \textit{Recall@k} metric.
On the other hand, models trained to classify candidate responses in a supervised fashion
make a significant improvement on the \textit{Recall@1} score
by going from $29\%$ with the baseline model and \emph{Rule-Based} selection, to $39\%$ with the more flexible \emph{Argmax} selection mechanism.

\subsection{Limitations}
\label{subsec:limitations}

Being partly motivated by an organized competition, some time constraints did not allow us to always pursue all the experiments planned.

The first set of limitations comes with the two \emph{HRED} models described in Section~\ref{subsubseq:hred}.
One extension could be to not only condition the decoder on the conversation history, but also on the news article's paragraph after being processed by a recurrent neural network.
This is especially true for the \emph{Reddit HRED} model that can retrieve the online article that triggered the \emph{Reddit} conversation.
In addition, instead of vanilla \emph{HRED} models, adding an \emph{Attention} mechanism \citep{bahdanau2014attention} could help the models generate less generic responses.

Another model that could have been improved is the \emph{Topic Classifier} model (Section~\ref{subsubseq:topic}).
As of now, a simple 10-class classifier and some predefined sentences are used to inform the user about the general topic.
However, there is a lot of work done in the area of text summarizarion that could be explored. For cases in which the news paragraph is quite long, a summarization model could be beneficial.

Regarding the different scoring techniques, one limitation is that the deep architecture involving Gated Recurrent Unit (GRU) networks, was not pre-trained on large corpora of text.
Training the recurrent networks to encode and decode conversational text (just like the \emph{HRED} models) could be beneficial for the scoring models.
Furthermore, different deep architectures may yield better results in the classification task presented in Section~\ref{subsubseq:sup_scorer}.


\subsection{Future Work}
\label{subseq:future_work}

Eventually, we leave as future work the task of training an end-to-end version of the presented system.
As previously mentioned, the nine generative systems producing candidate responses were trained once before the \emph{ConvAI} competition and remained fixed thereafter.
Thus, even if we had a \textit{Recall@1} score of 100\%, the system would still be limited by the capabilities of its components.

Finally, 
organizing academic competitions like the \emph{Amazon Alexa Prize} and the \emph{ConvAI} challenge are good alternatives to evaluate conversational agents in various tasks.
This work shows that making the data available can be useful to drive dialog research. 
Future challenges should thus encourage participation to have a good amount of evaluated conversational data, and release the data after the end of the competition.
As described in Appendix~\ref{subapp:tech_difficulties}, our team encountered several engineering difficulties in deploying the system in a live environment.
Often taking more time than expected, these challenges can reduce the amount of innovation in a system and discourage researchers from participating.
Future academic challenges should thus provide as much help as possible to deploy systems in an easy and secure fashion.

%% file: sections/appendix_1.tex
\section{Technical Details of the proposed system}
\label{app:appendix_tech_details}

\subsection{Challenge requirements}
\label{subapp:challenge_req}

\begin{figure}[ht!]
  \centering
  \resizebox{9cm}{!}{\input{fig/system_fig.tikz}}
  \caption{Overall system framework}
  \label{fig:system_fig}
\end{figure}
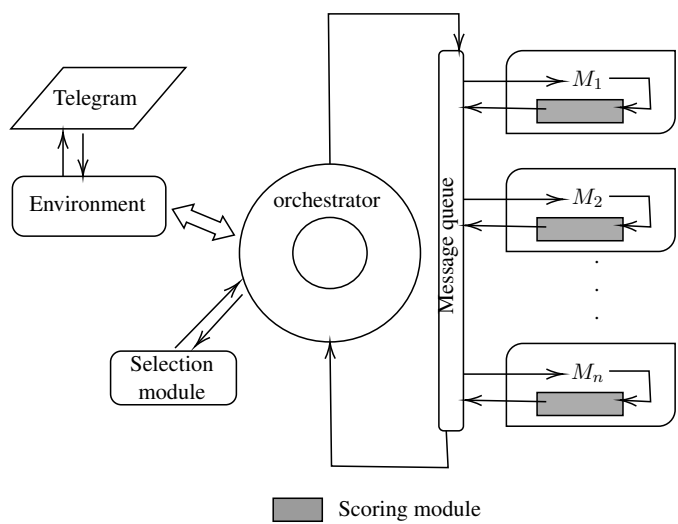

The requirements of the \textit{Conversational AI (ConvAI)} competition was to submit a self-contained model in a Docker \footnote{\href{https://www.docker.com/}{https://www.docker.com/}} instance. The competition environment used Telegram messaging platform to pair bots with human users. Since a ranker mechanism chooses the best response from an ensemble of models, \textit{all} models need to be loaded into memory at inference time. Individual models require variable amount of system memory, from the highest being the generative models and the lowest being the rule-based systems. Thus, a multiprocessing \textit{orchestrator} communicating via inter-process communication (IPC) message queues is implemented (Figure \ref{fig:system_fig}).

\subsection{Overall framework}
\label{subapp:framework}

\noindent At first, the \textit{orchestrator} receives a start-of-conversation signal from the environment (the ConvAI framework talking to Telegram), followed by a randomly assigned news article paragraph.
Then, the \textit{orchestrator} fires a \textit{wake-up} initialization call containing the news paragraph to all its child processes. Each of them contain one model from the ensemble.
The models themselves can then choose to initiate
themselves
with the paragraph text.
For example, the \textit{Entity Sentences} model (Section~\ref{subsubseq:entity_sents}) runs the \emph{Spacy Named Entity Recognizer}\footnote{\href{https://spacy.io/api/annotation}{https://spacy.io/api/annotation}} to pre-select a set of entities; the \textit{Topic classifier} model (Section~\ref{subsubseq:topic}) runs the \emph{fastText} classifier \citep{joulin2016fasttext} over the article text and saves the topic in its own dictionary.
Subsequently, the \textit{orchestrator} fires the question generator models (\textit{Neural Question Generator (NQG)} (Section~\ref{subsubseq:nqg}) and \textit{Entity Sentences}) to start proactively conversing about the given article.
Thereafter, on each turn, the \textit{orchestrator} shares the user response to all the child processes who generate their individual responses and submit them to the
\textit{scoring} module.
As soon as a candidate response is being generated, the \textit{scoring} module evaluates it and submits the response along with its score to the message queue bus.
Eventually, the \textit{selection} mechanism selects the best response based on its score.

\subsection{Message response latency}
\label{subapp:msg_response_latency}

One of the critical constraints of the setup is to reduce message response latency so that human judges do not have to wait more than a few seconds to respond to the query. In this setup, the wait time can be potentially compounded depending on individual response generation and scoring times. Therefore, the score of each response is calculated \textit{within} the model's own child process. A hard wait time of 7 seconds is set for the \textit{orchestrator} to listen to the message queue bus and it rejects any responses which arrive too late for processing.
When candidate responses arrive at the selection module, the scores are already computed, thus the module has only to sample accordingly and return one response to the \textit{orchestrator}.
Furthermore, since multiple users can communicate at the same time with the system, an incoming/outgoing message queue is implemented in the \textit{orchestrator} so that each model and the \textit{orchestrator} can communicate asynchronously through IPC.

\subsection{Choice of Inter-process communication system}
\label{subapp:inter-process_communication}

ZMQ \footnote{\href{http://zeromq.org/bindings:python}{http://zeromq.org/bindings:python}} was first explored to have an individual incoming/outgoing thread within each model child process. ZMQ provides an IPC message bus highway for fast communication. However, due to system limitations, spawning two threads (input and output) per child process (per model) increases the complexity of the overall system and results in increased response latency. We thereafter switched to Python's inbuilt shared message queue to reduce system complexity.
However, production systems having unconstrained system requirements would benefit from using a dedicated message queue such as ZMQ or ActiveMQ \footnote{\href{http://activemq.apache.org/}{http://activemq.apache.org/}}.

\subsection{Monitoring child processes}
\label{subapp:child_process}

Since each model runs on its own child process, a fail-safe redundant process monitoring system is implemented to handle the possibility of a model crash. The \textit{orchestrator} pings each model at every interaction, and if one model fails to respond within 60 seconds, then the \textit{orchestrator} revive the child process of that specific model.
This helps to have as many candidate responses as possible for each interaction.

\subsection{Technical Difficulties}
\label{subapp:tech_difficulties}


Part of the challenge was the engineering effort of exporting a research project into a real system with all the constraints that comes with it such as latency, concurrency, memory, and others.
For instance, one instruction received at the end of the challenge was that the submitted systems must have the following hardware constraints: 2 virtual processors Intel Xeon CPU @ 2.40GHz, 16 Gb of RAM, and 50 Gb of disk space. This proved to be critical for the system as the \emph{RLLChatbot} took about 50 Gb of RAM to load all the models in memory. This bottleneck resulted in increased latency of response due to operating system memory swapping and context switching.
Future improvements can be made by reducing the model size by using mixed precision systems \footnote{\href{https://docs.nvidia.com/deeplearning/sdk/mixed-precision-training/index.html}{https://docs.nvidia.com/deeplearning/sdk/mixed-precision-training/index.html}}.





%% file: fig/system_fig.tikz
\tikzset{every picture/.style={line width=0.75pt}} 

\begin{tikzpicture}[x=0.75pt,y=0.75pt,yscale=-1,xscale=1]

\draw   (463,62.8) .. controls (463,56.84) and (467.84,52) .. (473.8,52) -- (573,52) .. controls (573,52) and (573,52) .. (573,52) -- (573,95.2) .. controls (573,101.16) and (568.16,106) .. (562.2,106) -- (463,106) .. controls (463,106) and (463,106) .. (463,106) -- cycle ;
\draw  [fill={rgb, 255:red, 172; green, 172; blue, 172 }  ,fill opacity=1 ] (483,84) -- (539,84) -- (539,99) -- (483,99) -- cycle ;
\draw   (323.8,184) .. controls (323.8,171.19) and (334.63,160.8) .. (348,160.8) .. controls (361.37,160.8) and (372.2,171.19) .. (372.2,184) .. controls (372.2,196.81) and (361.37,207.2) .. (348,207.2) .. controls (334.63,207.2) and (323.8,196.81) .. (323.8,184)(289,184) .. controls (289,151.97) and (315.42,126) .. (348,126) .. controls (380.58,126) and (407,151.97) .. (407,184) .. controls (407,216.03) and (380.58,242) .. (348,242) .. controls (315.42,242) and (289,216.03) .. (289,184) ;
\draw   (205,255) .. controls (205,251.13) and (208.13,248) .. (212,248) -- (280,248) .. controls (283.87,248) and (287,251.13) .. (287,255) -- (287,276) .. controls (287,279.87) and (283.87,283) .. (280,283) -- (212,283) .. controls (208.13,283) and (205,279.87) .. (205,276) -- cycle ;
\draw   (141,141.8) .. controls (141,137.49) and (144.49,134) .. (148.8,134) -- (233.2,134) .. controls (237.51,134) and (241,137.49) .. (241,141.8) -- (241,165.2) .. controls (241,169.51) and (237.51,173) .. (233.2,173) -- (148.8,173) .. controls (144.49,173) and (141,169.51) .. (141,165.2) -- cycle ;
\draw  [fill={rgb, 255:red, 155; green, 155; blue, 155 }  ,fill opacity=1 ] (311,344) -- (343,344) -- (343,359) -- (311,359) -- cycle ;
\draw   (245.35,155.53) -- (259.31,152.24) -- (257.7,156.12) -- (276.49,163.93) -- (278.1,160.05) -- (285.61,172.27) -- (271.65,175.57) -- (273.26,171.69) -- (254.47,163.88) -- (252.86,167.76) -- cycle ;
\draw   (419,55.2) .. controls (419,53.43) and (420.43,52) .. (422.2,52) -- (431.8,52) .. controls (433.57,52) and (435,53.43) .. (435,55.2) -- (435,296.8) .. controls (435,298.57) and (433.57,300) .. (431.8,300) -- (422.2,300) .. controls (420.43,300) and (419,298.57) .. (419,296.8) -- cycle ;
\draw    (435,72) -- (498,72) ;
\draw [shift={(500,72)}, rotate = 180] [color={rgb, 255:red, 0; green, 0; blue, 0 }  ][line width=0.75]    (10.93,-3.29) .. controls (6.95,-1.4) and (3.31,-0.3) .. (0,0) .. controls (3.31,0.3) and (6.95,1.4) .. (10.93,3.29)   ;

\draw    (489,90) -- (438,89.04) ;
\draw [shift={(436,89)}, rotate = 361.08000000000004] [color={rgb, 255:red, 0; green, 0; blue, 0 }  ][line width=0.75]    (10.93,-3.29) .. controls (6.95,-1.4) and (3.31,-0.3) .. (0,0) .. controls (3.31,0.3) and (6.95,1.4) .. (10.93,3.29)   ;

\draw    (291,211) -- (261.35,243.52) ;
\draw [shift={(260,245)}, rotate = 312.36] [color={rgb, 255:red, 0; green, 0; blue, 0 }  ][line width=0.75]    (10.93,-3.29) .. controls (6.95,-1.4) and (3.31,-0.3) .. (0,0) .. controls (3.31,0.3) and (6.95,1.4) .. (10.93,3.29)   ;

\draw    (247,246) -- (287.6,204.43) ;
\draw [shift={(289,203)}, rotate = 494.33] [color={rgb, 255:red, 0; green, 0; blue, 0 }  ][line width=0.75]    (10.93,-3.29) .. controls (6.95,-1.4) and (3.31,-0.3) .. (0,0) .. controls (3.31,0.3) and (6.95,1.4) .. (10.93,3.29)   ;

\draw    (348,126) -- (347,28) -- (432,28) -- (432.18,50) ;
\draw [shift={(432.2,52)}, rotate = 269.52] [color={rgb, 255:red, 0; green, 0; blue, 0 }  ][line width=0.75]    (10.93,-3.29) .. controls (6.95,-1.4) and (3.31,-0.3) .. (0,0) .. controls (3.31,0.3) and (6.95,1.4) .. (10.93,3.29)   ;

\draw    (423.8,300) -- (425,324) -- (349,323) -- (348.02,244) ;
\draw [shift={(348,242)}, rotate = 449.29] [color={rgb, 255:red, 0; green, 0; blue, 0 }  ][line width=0.75]    (10.93,-3.29) .. controls (6.95,-1.4) and (3.31,-0.3) .. (0,0) .. controls (3.31,0.3) and (6.95,1.4) .. (10.93,3.29)   ;

\draw    (530,70) -- (556,70) -- (557,90) -- (539,89.1) ;
\draw [shift={(537,89)}, rotate = 362.86] [color={rgb, 255:red, 0; green, 0; blue, 0 }  ][line width=0.75]    (10.93,-3.29) .. controls (6.95,-1.4) and (3.31,-0.3) .. (0,0) .. controls (3.31,0.3) and (6.95,1.4) .. (10.93,3.29)   ;

\draw   (463,139.8) .. controls (463,133.84) and (467.84,129) .. (473.8,129) -- (573,129) .. controls (573,129) and (573,129) .. (573,129) -- (573,172.2) .. controls (573,178.16) and (568.16,183) .. (562.2,183) -- (463,183) .. controls (463,183) and (463,183) .. (463,183) -- cycle ;
\draw  [fill={rgb, 255:red, 172; green, 172; blue, 172 }  ,fill opacity=1 ] (483,161) -- (539,161) -- (539,176) -- (483,176) -- cycle ;
\draw    (435,149) -- (498,149) ;
\draw [shift={(500,149)}, rotate = 180] [color={rgb, 255:red, 0; green, 0; blue, 0 }  ][line width=0.75]    (10.93,-3.29) .. controls (6.95,-1.4) and (3.31,-0.3) .. (0,0) .. controls (3.31,0.3) and (6.95,1.4) .. (10.93,3.29)   ;

\draw    (489,167) -- (438,166.04) ;
\draw [shift={(436,166)}, rotate = 361.08000000000004] [color={rgb, 255:red, 0; green, 0; blue, 0 }  ][line width=0.75]    (10.93,-3.29) .. controls (6.95,-1.4) and (3.31,-0.3) .. (0,0) .. controls (3.31,0.3) and (6.95,1.4) .. (10.93,3.29)   ;

\draw    (530,147) -- (556,147) -- (557,167) -- (539,166.1) ;
\draw [shift={(537,166)}, rotate = 362.86] [color={rgb, 255:red, 0; green, 0; blue, 0 }  ][line width=0.75]    (10.93,-3.29) .. controls (6.95,-1.4) and (3.31,-0.3) .. (0,0) .. controls (3.31,0.3) and (6.95,1.4) .. (10.93,3.29)   ;

\draw   (463,253.8) .. controls (463,247.84) and (467.84,243) .. (473.8,243) -- (573,243) .. controls (573,243) and (573,243) .. (573,243) -- (573,286.2) .. controls (573,292.16) and (568.16,297) .. (562.2,297) -- (463,297) .. controls (463,297) and (463,297) .. (463,297) -- cycle ;
\draw  [fill={rgb, 255:red, 172; green, 172; blue, 172 }  ,fill opacity=1 ] (483,275) -- (539,275) -- (539,290) -- (483,290) -- cycle ;
\draw    (435,263) -- (498,263) ;
\draw [shift={(500,263)}, rotate = 180] [color={rgb, 255:red, 0; green, 0; blue, 0 }  ][line width=0.75]    (10.93,-3.29) .. controls (6.95,-1.4) and (3.31,-0.3) .. (0,0) .. controls (3.31,0.3) and (6.95,1.4) .. (10.93,3.29)   ;

\draw    (489,281) -- (438,280.04) ;
\draw [shift={(436,280)}, rotate = 361.08000000000004] [color={rgb, 255:red, 0; green, 0; blue, 0 }  ][line width=0.75]    (10.93,-3.29) .. controls (6.95,-1.4) and (3.31,-0.3) .. (0,0) .. controls (3.31,0.3) and (6.95,1.4) .. (10.93,3.29)   ;

\draw    (530,261) -- (556,261) -- (557,281) -- (539,280.1) ;
\draw [shift={(537,280)}, rotate = 362.86] [color={rgb, 255:red, 0; green, 0; blue, 0 }  ][line width=0.75]    (10.93,-3.29) .. controls (6.95,-1.4) and (3.31,-0.3) .. (0,0) .. controls (3.31,0.3) and (6.95,1.4) .. (10.93,3.29)   ;

\draw   (174.2,63) -- (254,63) -- (219.8,103) -- (140,103) -- cycle ;
\draw    (174,134) -- (174,107) ;
\draw [shift={(174,105)}, rotate = 450] [color={rgb, 255:red, 0; green, 0; blue, 0 }  ][line width=0.75]    (10.93,-3.29) .. controls (6.95,-1.4) and (3.31,-0.3) .. (0,0) .. controls (3.31,0.3) and (6.95,1.4) .. (10.93,3.29)   ;

\draw    (187,106) -- (187,132) ;
\draw [shift={(187,134)}, rotate = 270] [color={rgb, 255:red, 0; green, 0; blue, 0 }  ][line width=0.75]    (10.93,-3.29) .. controls (6.95,-1.4) and (3.31,-0.3) .. (0,0) .. controls (3.31,0.3) and (6.95,1.4) .. (10.93,3.29)   ;

\draw (245,265) node  [align=center] {Selection\\module};
\draw (346,149) node  [align=left] {orchestrator};
\draw (516,71) node   {$M_{1}$};
\draw (190,152) node  [align=left] {Environment};
\draw (400,352) node  [align=left] {Scoring module};
\draw (425,179) node [rotate=-270] [align=left] {Message queue};
\draw (522,208) node  [align=left] {.\\.\\.};
\draw (516,148) node   {$M_{2}$};
\draw (516,262) node   {$M_{n}$};
\draw (195,84) node  [align=left] {Telegram};

\end{tikzpicture}

%% file: sections/appendix_2.tex
\section{Hyperparameter \& Implementation details of experiments}
\label{app:hyperparams}

\subsection{Explored parameters}
\label{subapp:param_explored}

For all experiments described in Section~\ref{subseq:experiments}, different combinations of the following parameters are explored by randomly sampling 100 values:
\begin{itemize}[noitemsep,topsep=0pt]
    \item optimizer: \emph{ADAM} \citep{kingma2014adam}, \emph{SGD} \citep{rumelhart1985learning,rumelhart1986learning}, \emph{Adagrad} \citep{duchi2011adagrad}, \emph{Adadelta} \citep{zeiler2012adadelta}, and \emph{RMSProp} \citep{hinton2012rmsprop}.
    \item learning rate: $0.01$, $0.001$, and $0.0001$.
    \item activation function: \emph{Sigmoid}, \emph{ReLU} \citep{glorot2011relu}, and \emph{pReLU} \citep{he2015kaiming_init}.
    \item weight initialization: \emph{He} \citep{he2015kaiming_init}, and \emph{Glorot} \citep{glorot2010difficultTrainFFNN}
    \item dropout rate: $0.2$, $0.4$, $0.6$, and $0.8$
\end{itemize}
These are the only flexible parameters in order to limit the number of degrees of freedom in the system.
Moreover, these parameters are expected to have a direct influence on the training behavior of the system. 
The architecture of the networks is kept fixed because the size of the networks only influence the capacity of the models rather than their ability to learn.

\subsection{Best \{Small/Deep\}\_R parameters}
\label{subapp:best_r_params}

After running 100 \emph{SmallR} experiments and another 100 \emph{DeepR} experiments with different random parameter combinations as described in Appendix~\ref{subapp:param_explored}, the \emph{SmallR} and \emph{DeepR} models with highest F1 validation score were trained with:
\begin{itemize}[noitemsep,topsep=0pt]
    \item a batch size of $128$,
    \item the \emph{RmsProp} and \emph{SGD} optimizers respectively,
    \item a learning rate of $0.0001$ and $0.001$ respectively,
    \item \emph{pReLU} activation functions with \emph{He} weight initialization \citep{he2015kaiming_init},
    \item and a dropout rate of $0.2$ and $0.4$ respectively.
\end{itemize}
This combination of parameters gave a validation F1 score of $0.420$ for the best \emph{SmallR} model, and a validation F1 score of $0.373$ for the best \emph{DeepR} model.

\subsection{Best \{Small/Deep\}\_Q parameters}
\label{subapp:best_q_params}

After running 100 \emph{SmallQ} experiments and another 100 \emph{DeepQ} experiments with different random parameter combinations as described in Appendix~\ref{subapp:param_explored}, the \emph{SmallQ} and \emph{DeepQ} models with lowest validation loss were trained with:
\begin{itemize}[noitemsep,topsep=0pt]
  \item the regular training set (i.e.: not the over-sampled one),
  \item a batch size of $128$,
  \item the \emph{SGD} and \emph{ADAM} optimizers respectively,
  \item a learning rate of $0.0001$,
  \item a discount factor of $0.99$,
  \item an update frequency of $2,000$ updates for the target DQN,
  \item a hidden size of $300$ for the recurrent networks in \emph{DeepQ} experiments,
  \item \emph{sigmoid} activation functions with \emph{Glorot} weight initialization \citep{glorot2010difficultTrainFFNN},
  \item and a dropout rate of $0.8$.
\end{itemize}
This combination of parameters gave a minimal validation loss of $0.00488$ for the best \emph{SmallQ} model, and a minimal validation loss of $0.00505$ for the best \emph{DeepQ} model.